%% file: arxiv.tex
\documentclass[10pt,twocolumn,letterpaper]{article}

\usepackage{iccv}
\usepackage{times}
\usepackage{epsfig}
\usepackage{graphicx}
\usepackage{amsmath}
\usepackage{amssymb}

\usepackage[breaklinks=true,bookmarks=false, colorlinks]{hyperref}

\newcommand{\modelname}{Video-$k$MaX\xspace}

\newcommand{\methodname}{HiLA-MB\xspace}

\include{math_commands}

\newcommand{\methodnamefull}{Hierarchical Location-Aware Memory Buffer\xspace}
\iccvfinalcopy 

\definecolor{plotsgreen}{RGB}{38,150,38}
\definecolor{plotsorange}{RGB}{255,115,17}
\definecolor{plotsred}{RGB}{208,34,35}
\definecolor{plotsyellow}{RGB}{255,225,80}
\definecolor{plotsgrey}{RGB}{128,128,128}
\definecolor{plotspurple}{RGB}{137,92,181}
\definecolor{plotsblue}{RGB}{0,101,167}

\usepackage{caption}
\usepackage{subcaption}

\newlength\savewidth\newcommand\shline{\noalign{\global\savewidth\arrayrulewidth
  \global\arrayrulewidth 1pt}\hline\noalign{\global\arrayrulewidth\savewidth}}

\usepackage{xcolor,colortbl}
\usepackage{multirow}
\def\iccvPaperID{2068} 

\usepackage{tikz}
\usetikzlibrary{arrows,intersections}
\tikzset{font=\scriptsize}
\usepackage{pgfplots}
\pgfplotsset{compat=1.11}
\usepgfplotslibrary{fillbetween}
\usetikzlibrary{backgrounds}

\usepackage[capitalize]{cleveref}
\crefname{section}{Sec.}{Secs.}
\Crefname{section}{Section}{Sections}
\Crefname{table}{Table}{Tables}
\crefname{table}{Tab.}{Tabs.}

\newcommand{\methodbaselinefull}{na\"ive Memory Buffer\xspace}
\newcommand{\methodbaseline}{na\"ive-MB\xspace}
\newcommand{\figref}[1]{Fig\onedot~\ref{#1}}
\newcommand{\equref}[1]{Eq\onedot~\eqref{#1}}
\newcommand{\secref}[1]{Sec\onedot~\ref{#1}}
\newcommand{\tabref}[1]{Tab\onedot~\ref{#1}}

\definecolor{baselinecolor}{rgb}{0.94, 0.8, 0.62}
\newcommand{\baseline}[1]{\cellcolor{baselinecolor}{#1}}

\definecolor{settingcolor}{gray}{0.9}
\newcommand{\setting}[1]{\cellcolor{settingcolor}{#1}}

\definecolor{lightblue}{rgb}{0.85,0.85,1}
\newcommand{\vsbaseline}[1]{\cellcolor{lightblue}{#1}}

\definecolor{lightred}{rgb}{0.96,0.71,0.7}
\newcommand{\idolbaseline}[1]{\cellcolor{lightred}{#1}}

\makeatletter
\newcommand{\thickhline}{%
    \noalign {\ifnum 0=`}\fi \hrule height 1pt
    \futurelet \reserved@a \@xhline
}

\ificcvfinal\fi

\begin{document}
\title{Video-kMaX: A Simple Unified Approach for Online and Near-Online \\ Video Panoptic Segmentation}

\author{
Inkyu Shin\textsuperscript{1,2$\dagger$}~~~~Dahun Kim\textsuperscript{2}~~~~~~~~Qihang Yu\textsuperscript{$\dagger$}~~~~Jun Xie\textsuperscript{2}~~~~Hong-Seok 
Kim\textsuperscript{2}~~~Bradley Green\textsuperscript{2}\\
In So Kweon\textsuperscript{1}~~~~Kuk-Jin Yoon\textsuperscript{1}~~~~~~Liang-Chieh Chen\textsuperscript{$\dagger$}\\\textsuperscript{1}KAIST~~~~~\textsuperscript{2}Google Research
}

\maketitle

\input{sections/0.abstract}
\input{sections/1.intro}
\input{sections/2.related}

\input{sections/3.method}
\input{sections/4.experiments}
\input{sections/5.conclusion}

\ificcvfinal\thispagestyle{empty}\fi

{\small
\bibliographystyle{ieee_fullname}
\bibliography{egbib}
}

\end{document}

%% file: math_commands.tex
\renewcommand{\Re}{\mathbb{R}}

\renewcommand{\Sigma}{\mathfrak{S}}

\def\1{\bm{1}}

\DeclareMathAlphabet{\mathsfit}{\encodingdefault}{\sfdefault}{m}{sl}
\SetMathAlphabet{\mathsfit}{bold}{\encodingdefault}{\sfdefault}{bx}{n}

\DeclareMathOperator*{\argmax}{arg\,max}

%% file: sections/0.abstract.tex
\begin{abstract}
    Video Panoptic Segmentation (VPS) aims to achieve comprehensive pixel-level scene understanding by segmenting all pixels and associating objects in a video.
    Current solutions can be categorized into online and near-online approaches.
    Evolving over the time, each category has its own specialized designs, making it nontrivial to adapt models between different categories.
    To alleviate the discrepancy, in this work, we propose a unified approach for online and near-online VPS.
    The meta architecture of the proposed \modelname consists of two  components: within-clip segmenter (for clip-level segmentation) and cross-clip associater (for association beyond clips).
    We propose clip-$k$MaX (clip $k$-means mask transformer) and \methodname (\methodnamefull) to instantiate the segmenter and associater, respectively.
    Our general formulation includes the online scenario as a special case by adopting clip length of one.
    Without bells and whistles, \modelname sets a new state-of-the-art on KITTI-STEP and VIPSeg for video panoptic segmentation, and VSPW for video semantic segmentation.
    Code will be made publicly available.
    \let\thefootnote\relax\footnote{$^\dagger$Work done while at Google.}
\end{abstract}

%% file: sections/1.intro.tex
\section{Introduction}
\label{sec:intro}
Video Panoptic Segmentation (VPS)~\cite{kim2020video} aims at a holistic video understanding of the scene by unifying two  challenging tasks: semantically segmenting images and associating segmented regions across all frames in a video~\cite{weber2021step}. 
It can benefit various real-world applications, such as autonomous driving, robot visual control, and video editing. 

\begin{figure}
\begin{center}
\includegraphics[width=1.00\linewidth]{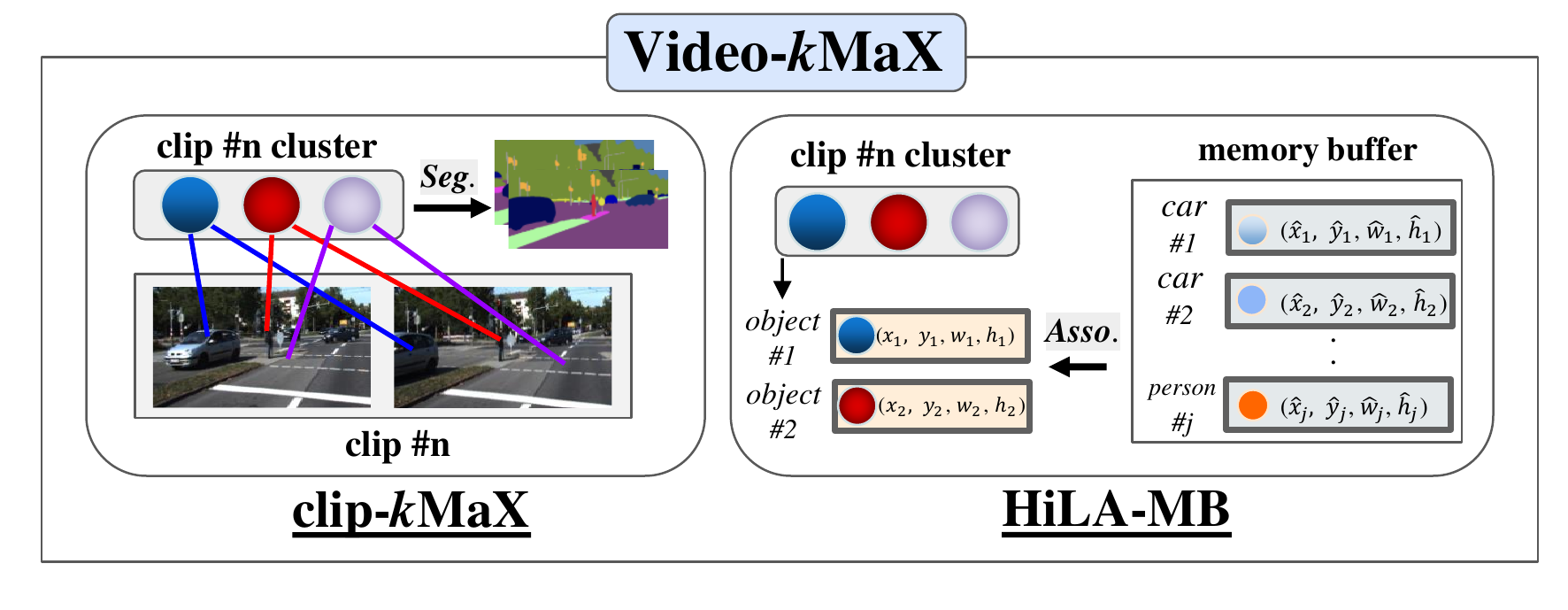}
\caption{
The meta architecture of \modelname consists of clip-$k$MaX for clip-level segmentation and \methodname for object association. The former groups pixels of the same object within-clip and the latter leverages appearance and location features (encoded by box coordinates) for long-term association across-clips.
}
\label{fig:teaser}
\end{center}
\vspace{-6mm}
\end{figure}


\begin{figure}
\begin{center}
\includegraphics[width=0.98\linewidth]{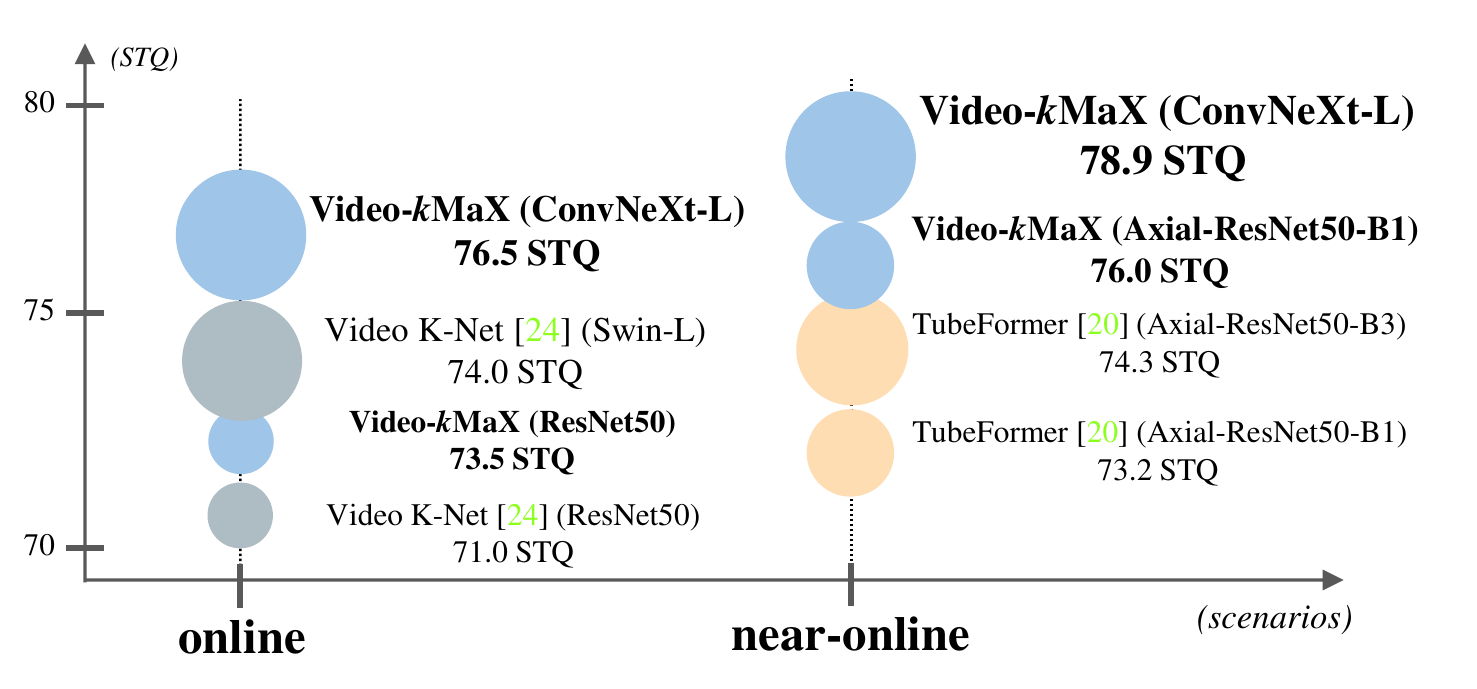}
\caption{
\modelname is a unified approach for online and near-online video panoptic segmentation, showing state-of-the-art performance in both scenarios (on KITTI-STEP \textit{val} set).
The size of the circle reflects the model parameters.
}
\label{fig:score}
\end{center}
\end{figure}

With the rapid growth of interest, there have been several methods~\cite{kim2020video, woo2021learning, vip_deeplab, kim2022tubeformer, li2022video, zhou2022slot} proposed for VPS.
They can be categorized into online and near-online approaches, which process the video either frame-by-frame or clip-by-clip (a clip contains only a few consecutive video
frames).
The online approaches, such as VPSNet~\cite{kim2020video} and Video K-Net~\cite{li2022video}, segment each frame sequentially via the modern image-level segmenter~\cite{he2017mask, xiong19upsnet, zhang2021k}, and build an additional association branch trained to enforce consistent predictions between frames~\cite{Yang19ICCV, li2022video}.
On the other hand, the near-online approaches, such as ViP-DeepLab~\cite{vip_deeplab} and TubeFormer~\cite{kim2022tubeformer}, extend the modern image-level segmenter~\cite{cheng2020panoptic, wang2021max} to process a clip by designing extra modules (\eg, next-frame instance segmentation~\cite{vip_deeplab} or latent memory~\cite{kim2022tubeformer}). The clip-level predictions are then stitched~\cite{vip_deeplab} to form the final video segmentation results.
The modules designed for online or near-online approaches are not only evolving over time, but also becoming very distinct for each scenario.
Consequently, it is infeasible to easily adapt online models to near-online (and vice versa).
Particularly, the current online methods~\cite{kim2020video,li2022video} lack a proper clip-level segmenter, while the modern near-online methods~\cite{vip_deeplab,kim2022tubeformer} fail to associate objects in an online manner, suffering from the absence of overlapping frames. 
The need for this scenario-specific design results in inefficiencies, as it requires the utilization of different frameworks for each setting.
A natural question thus emerges:
\textit{Is it possible to develop a unified framework for online and near-online VPS without any scenario-specific design?}

To answer the question, we carefully design \modelname, a simple yet effective approach for both online and near-online VPS.
As drawn in~\figref{fig:teaser}, the meta architecture of \modelname contains two components: within-clip segmenter and cross-clip associater, where the former component performs clip-level segmentation and the later one associates detected objects across clips.
The proposed \modelname is an instantiation of the pipeline by adopting clip-$k$MaX (clip $k$-means mask transformer) for the within-clip segmenter, and \methodname (\methodnamefull) for the cross-clip associater.

The proposed clip-$k$MaX extends the image-level $k$-means mask transformer~\cite{yu2022k} to the clip-level \textit{without} adding any extra modules or loss functions.
Motivated by the $k$-means clustering perspective~\cite{yu2022cmt}, we consider object queries as cluster centers, where each query is responsible for grouping pixels of the same object \textit{within} a clip together.
Specifically, each object query, when multiplied with the clip features~\cite{tian2020conditional,wang2020solov2,wang2021max}, is learned to yield a \textit{tube} prediction (\ie, masks of the same object in a clip)~\cite{kim2022tubeformer}.
This learning can be achieved via a surprisingly simple modification in the $k$-means cross-attention module~\cite{yu2022k} by concatenating the clip-level pixel features along the spatial dimension.
As a result, clip-$k$MaX can be applied to both near-online and online settings without additional complexities.
We also empirically show that $k$-means cross-attention is an effective mechanism for handling the extremely long sequence of spatially and temporally flattened clip features. 

The proposed \methodname is motivated by the drawbacks of existing methods through the careful systematical studies.
We observe that the modern VPS methods~\cite{kim2020video, vip_deeplab} could not handle the more challenging setting of \textit{long-term} object tracking, since they either associate objects in the neighboring frames~\cite{kim2020video} or stitch overlapping frame predictions~\cite{vip_deeplab}, making it hard to track objects beyond the short clip length.
One promising solution is to exploit a memory buffer to propagate the tracking information across all video clips, which has been proven successful in the recent works~\cite{Yang19ICCV,zeng2022motr,wu2022defense,huang2022minvis}.
However, surprisingly, we observe that na\"ively adopting the memory buffer to VPS leads to minor improvements or even worse performance.
The setback enforces us to further look into its root case.
We discover that the appearance feature alone~\cite{kim2020video,wu2022defense} is not sufficient for long-term association in VPS, when the target object is occluded for a long time; additionally, the memory buffer approach accumulates many detected objects, resulting in a huge matching space (between stored and newly detected objects) and hindering the matching accuracy.
To resolve the issues, we develop \methodname (\methodnamefull), which effectively incorporates location information to the memory module by two means.
First, when comparing the similarity between the stored objects in memory and the detected objects in the current frame, we consider not only their appearance features (encoded by object queries), but also their location features (encoded by normalized bounding box coordinates).
Specifically, if the object of interest is not detected in the current frame but it is stored in the memory (\eg, due to occlusion), we will ``predict" its current location by assuming the object is moving at a constant velocity.
Second, we propose a \textit{hierarchical} matching scheme to effectively reduce the matching space.
We initially exploit the matching results from the Video Stitching~\cite{vip_deeplab} strategy, which associates objects based on their mask IoU in the overlapping frame between clips, effective for short-term association.
We then associate the objects stored in memory with the currently detected but \textit{unmatched} objects, aiming for long-term association.
Thanks to our careful design, the \methodname improves the long-term association quality both in near-online and online scenarios with low sensitivity to the hyper-parameter values.


In summary, we introduce \modelname, a simple and unified method for online and near-online VPS. 
Our approach, consisting of two seamless modules: clip-$k$MaX and \methodname, achieves significant performance improvements on two long sequence VPS datasets: KITTI-STEP~\cite{weber2021step} and VIPSeg\cite{miao2022large}.
In particular, as shown in~\figref{fig:score}, our best \modelname outperforms the previous state-of-the-art online model (Video K-Net~\cite{li2022video}) and near-online model (TubeFormer~\cite{kim2022tubeformer}) by \textbf{+2.5\%} STQ and \textbf{+4.6\%} STQ, respectively, on KITTI-STEP val set.
We also show that our \modelname is scalable to another task, video semantic segmentation with VSPW~\cite{miao2021vspw} dataset by outperforming the baselines.

%% file: sections/2.related.tex
\begin{figure*}[!t]
\begin{center}
\includegraphics[width=0.8\linewidth]{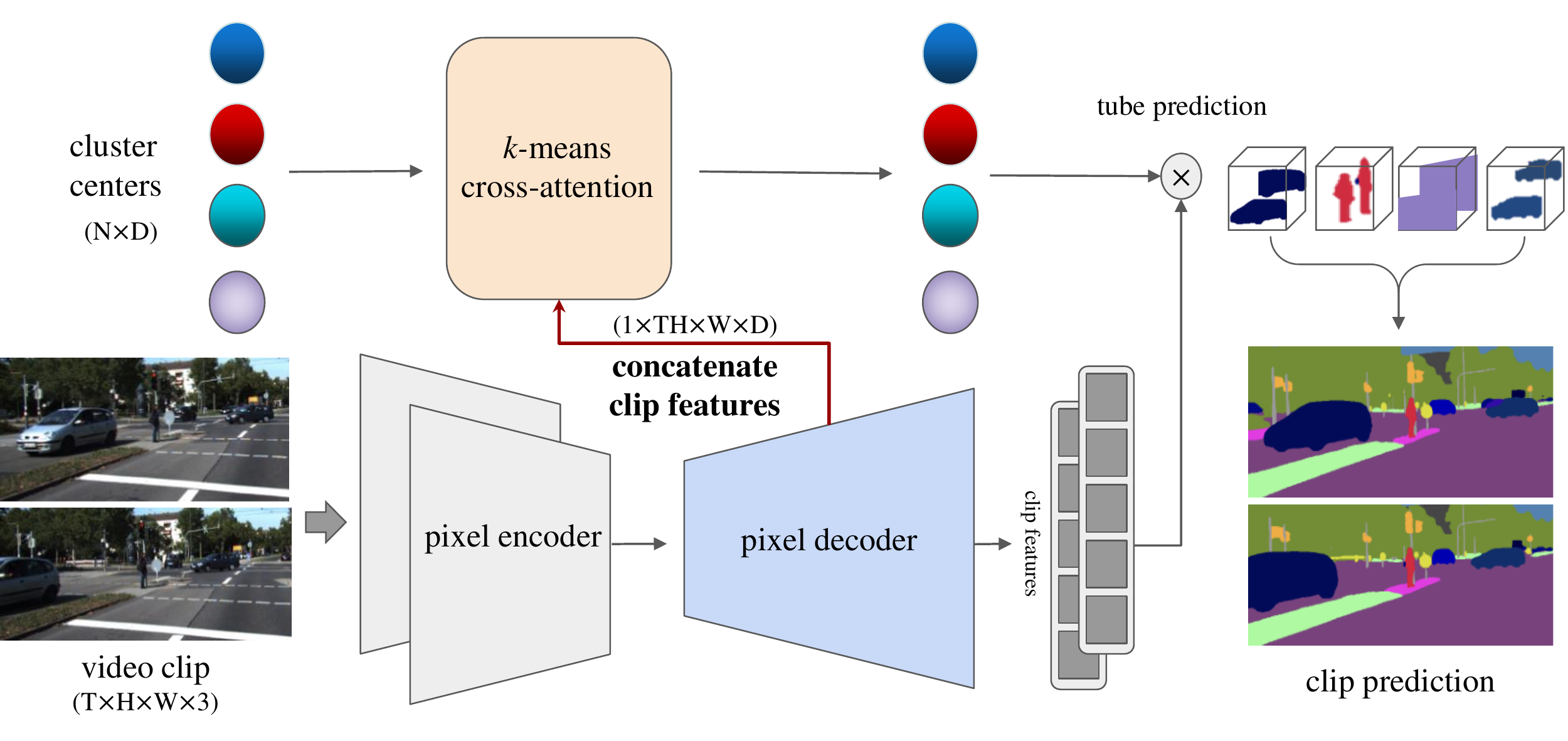}
\caption{{\bf Illustration of clip-$k$MaX.} The proposed clip-$k$MaX seamlessly converts the image-level segmentation model $k$MaX-DeepLab to clip-level \textit{without} adding extra module.
Motivated by the $k$-means perspective, clip-$k$MaX considers one object query as one cluster center, which learns to group together pixels of the same object \textit{within} a clip.
Specifically, each object query, when multiplied with the clip features, is learned to generate a \textit{tube} prediction (\ie, masks of the same object in the clip).
This learning can be accomplished by a surprisingly simple change in the $k$-means cross-attention module by concatenating the clip-level pixel features along the spatial dimension (\ie, treating the clip-level pixel features with shape $T\times H \times W \times D$ as one large image-level pixel feature with shape $1\times TH \times W \times D$), where the input video clip contains $T$ frames with height $H$ and width $W$. $D$ is the channel dimension of pixel features, and $N$ is number of cluster centers.
}
\label{fig:method_kmax}
\end{center}
\end{figure*}

\section{Related Work}
\label{sec:related}
{\bf Video Panoptic Segmentation (VPS)}~\quad
Video Panoptic Segmentation~\cite{kim2020video} aims to unify video semantic~\cite{zhu2017deep,gadde2017semantic,jain2019accel, miao2021vspw} and video instance~\cite{voigtlaender2019mots,Yang19ICCV, bertasius2020classifying,hwang2021video} segmentation.
Numerous efforts have been to transform image panoptic segmentation models~\cite{kirillov2018panoptic,kirillov2019panoptic,xiong2019upsnet,yang2019deeperlab,li2018attention,cheng2019panopticworkshop,wang2020axial,wang2021max,li2021fully} to the video domain. 
Among them, the online method VPSNet~\cite{kim2020video} adopts task-specific prediction heads from instance segmentation~\cite{he2017mask}, semantic segmentation~\cite{dai2017deformable}, and tracking~\cite{Yang19ICCV}, and jointly trains them to obtain panoptic video results. Similarly, the near-online method ViP-DeepLab~\cite{vip_deeplab} adds a next-frame instance segmentation head on top of Panoptic-DeepLab~\cite{cheng2020panoptic} that provides generic image panoptic segmentation with dual-ASPP~\cite{chen2017deeplab} semantic segmentation module and dual-decoder~\cite{chen2018encoder} based instance segmentation.
More recent works~\cite{kim2022tubeformer, zhou2022slot, li2022video} identifies the limitations of previous methods that require multiple separate networks and complex post-processing (\eg, NMS, fusion for tracking).
To address the issues, they design a transformer architecture~\cite{carion2020end} for end-to-end video panoptic segmentation.
However, all these methods have two fundamental issues.
First, they require specific designs for either online or near-online scenario, \eg, another association module~\cite{kim2020video,li2022video}, temporal consistency loss~\cite{kim2022tubeformer,li2022video}, or clip-segmentation module~\cite{vip_deeplab,kim2022tubeformer}.
Second, the models could only deal with short-term association (\ie, either neighboring frames or a clip).
In this regard, we propose a simple unified online and near-online video panoptic  segmentation model for long-term association without adding extra scenario-specific designs. 

{\bf Memory Module for Long-Term Tracking}~\quad
Object queries from the Transformer decoder~\cite{carion2020end} have been used to track objects in multi-object tracking~\cite{sun2020transtrack,meinhardt2022trackformer,cai2022memot,zeng2022motr,zhao2022tracking}, video instance segmentation~\cite{hwang2021video,wu2022efficient,cheng2021mask2former,wu2022seqformer,wu2022defense,huang2022minvis,heo2022vita, heo2022generalized, UNINEXT}, and video panoptic segmentation~\cite{kim2022tubeformer,li2022video,zhou2022slot}.
Some of them exploit queries for the short-term association~\cite{sun2020transtrack,meinhardt2022trackformer}, while the others for the long-term association by additionally exploiting the memory buffer~\cite{zeng2022motr,wu2022defense, heo2022generalized}.
Particularly,
MOTR~\cite{zeng2022motr} proposes a set of track queries to model the tracked objects in the entire video.
MeMOT~\cite{cai2022memot} develops a spatio-temporal memory that stores a long range states of all tracked objects.
MaskTrack R-CNN~\cite{Yang19ICCV} employs a memory module to track detected objects. 
To make the association robust to challenging scenarios, such as heavy occlusion, IDOL~\cite{wu2022defense} proposes a temporally weighted softmax score for object matching. 
Along the same direction, we specialize the memory buffer approach for both online and near-online video panoptic segmentation models, and additionally develop an efficient hierarchical matching scheme.

%% file: sections/3.method.tex
\section{Method}
\label{sec:method}
The meta architecture of \modelname contains two components: clip-$k$MaX (clip $k$-means mask transformer) for within-clip segmentation (\secref{sec:nearonline_vps}) and \methodname (\methodnamefull) for cross-clip association (\secref{sec:naive_mb}).
We detail them below, starting from the near-online framework.
Our general formulation includes the online scenario by using clip length one (\secref{sec:online}).


\subsection{Within-Clip Segmenter: clip-$k$MaX}
\label{sec:nearonline_vps}
We first present the general formulation for image and video panoptic segmentation, before introducing our within-clip segmenter clip-$k$MaX, which performs clip-level segmentation with a short length $T$ (\eg, $T=2$).


{\bf General Formulation for Image and Video}~\quad
Recently, image panoptic segmentation has been reformulated as a simple set prediction powered by Transformer~\cite{vaswani2017attention}.
From the pioneering works (\eg, DETR~\cite{carion2020end} and MaX-DeepLab~\cite{wang2021max}) to the recent state-of-the-art methods (\eg, $k$MaX-DeepLab~\cite{yu2022k}),
panoptic predictions are designed to match the ground truth masks by segmenting image $I\in\Re^{H\times W\times 3}$ into a fixed-size set of $N$ class-labeled masks:
\begin{equation}
    \{\hat{y}_i\}_{i=1}^N = \{(\hat{m}_i, \hat{p}_i(c))\}_{i=1}^N ,
    \label{eqn:panoptic_formulation}
\end{equation}
where $\hat{m}_i \in {[0,1]}^{H\times W}$ and $\hat{p}_i(c)$ denote predicted mask and semantic class probability for the corresponding mask, respectively. Motivated by this, TubeFormer~\cite{kim2022tubeformer} extends this formulation into set prediction of class-labeled \textit{tubes}: $\{\hat{y}_i\}_{i=1}^N = \{(\hat{m}_i, \hat{p}_i(c))\}_{i=1}^N$, where $\hat{m}_i \in {[0,1]}^{T\times H\times W}$.
In this setting, $N$ object queries attend to the $T\times H \times W$ clip features, and predict $N$ tubes.
The prediction generalizes well for different values of $T$, since the positional embedding is only performed in the frame level, providing a useful structural prior that the same object in neighboring frames (assuming slow motion) will still be assigned by the same object query.
Given the generalizability, we are able to absorb the $T$-axis into the $H$-axis before feeding the clip features to transformer decoder.
Specifically, we propose to relax \cref{eqn:panoptic_formulation} into a more general form:
$\{\hat{y}_i\}_{i=1}^N = \{(\hat{m}_i, \hat{p}_i(c))\}_{i=1}^N$, where $\hat{m}_i \in {[0,1]}^{S\times W}$, $S$=$TH$, and $T\geq$1 (\ie, $S$ can change according to the different number of frames $T$).
By doing so, it allows us to easily extend an image panoptic segmentation model to the video domain (clip-level), as detailed below.


\begin{figure*}[!t]
\begin{center}
\includegraphics[width=0.95\linewidth]{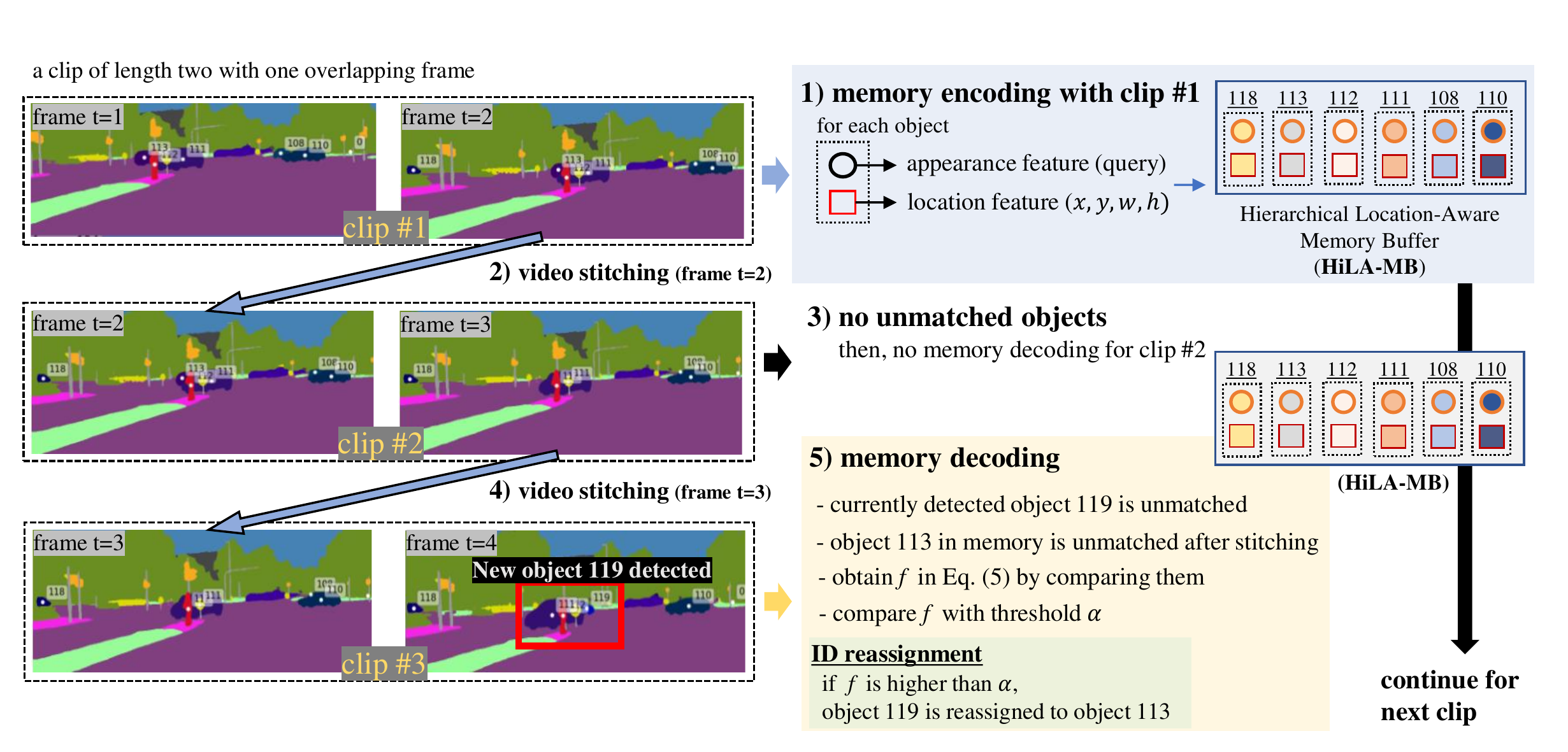}
\caption{{\bf Step-by-step overview of \methodnamefull (\methodname).}
The \methodname approach consists of two phases: Encoding and Decoding. In the Encoding Phase, appearance and location features of detected objects are stored in the memory buffer. In the Decoding Phase, \methodname performs hierarchical matching, beginning with video stitching for short-term association in overlapping frames between clips, followed by long-term association between unmatched objects in the current clip (\ie, object 119 in figure) and objects stored in memory (\ie, object 113 in figure).
}
\label{fig:method_lamb}
\end{center}
\end{figure*}

{\bf clip-$k$Max}~\quad
The state-of-the-art image segmentation model $k$MaX-DeepLab~\cite{yu2022k} replaces the cross-attention in a typical transformer decoder~\cite{vaswani2017attention} with $k$-means cross-attention by taking a cluster-wise argmax as below:

\begin{equation}
    \hat{C} = C + \argmax_N(Q^{c} \times (K^{p})^{T}) \times V^{p},
    \label{eqn:cross_attention}
\end{equation}
where $C\in\Re^{N \times D}$ refers to N object queries with D channels.
We use superscripts $p$ and $c$ to indicate the feature projected from the pixel features and object queries, respectively. $Q^c \in \mathbb{R}^{N \times D}, K^p \in \mathbb{R}^{HW \times D}, V^p \in \mathbb{R}^{HW \times D} $ stand for the linearly projected features for query, key, and value, respectively.
In this $k$-means perspective, one object query is regarded as a cluster center, which learns to group pixels of the same object together.
Given our previous general formulation, we can seamlessly extend $k$MaX-DeepLab to video clip, forming our clip-$k$MaX, by \textit{simply} reshaping the key and value into $K^{p}\in\Re^{SW\times D}$ and $V^{p}\in\Re^{SW\times D}$ ($S=TH$ and T$\geq$1).
The reshaping merges the $T$-frame feature to a \textit{single}-frame feature with large height $TH$ (\ie, reshape $T\times H \times W$ to $1 \times TH \times W$), which then becomes compatible with the image model $k$MaX-DeepLab. 
This is equivalent to performing the $k$-means clustering for a video clip with length $T$, where one query is now learning to group pixels of the same object \textit{in the clip} together.
We illustrate clip-$k$MaX in \figref{fig:method_kmax}.
Note that $k$MaX-DeepLab then becomes a special case of clip-$k$MaX with $T=1$.

{\bf Discussion}~\quad
The design of clip-$k$MaX may look simple on the surface. However, we made strenuous efforts in enhancing the conventional cross-attention module for clip-level mask predictions during its development.
When dealing with the extremely large sequence
length of spatially and temporally flattened clip features in a video clip, the standard cross-attention module is susceptible to learning, as each object query is required to identify the most distinguishable pixels among the abundant clip features.
This phenomenon was evident in the poor performance of the original cross-attention, motivating the prior art TubeFormer~\cite{kim2022tubeformer} to further employ an additional latent memory module.
To address this challenge, we propose using the $k$-means cross-attention~\cite{yu2022k} approach, which is capable of handling flattened clip features of any size by performing a cluster-wise argmax on $N$ cluster centers.
We will empirically prove this in the ablation studies.
{\bf Video Stitching (VS)}~\quad
In practice, given the limited memory, we are only able to perform clip-level inference (\ie, segmenting a short clip with length $T$).
To obtain the video-level segmentation, some heuristic designs are required.
One popular approach is Video Stitching (VS)~\cite{vip_deeplab,kim2022tubeformer}, which propagates object identities between clips by matching the mask IoU scores in the \textit{overlapping} frames.
In our framework, we adopt the same video stitching strategy for our near-online \modelname, but additionally explore memory buffer for long-term association. 




\subsection{Cross-Clip Associater: \methodname}
\label{sec:naive_mb}
 Our \methodname basically consists of two phases: Encoding Phase to store the previous object features, and Decoding Phase to associate current objects with the objects stored in the memory buffer. We detail the process below.

{\bf Encoding Phase}~\quad
The memory buffer is initially empty, when a new testing video comes.
It encodes features from all detected objects, while processing frames sequentially.
Regarding the object features to be stored, we exploit the appearance and location properties of each object.

For the appearance feature of object $i$ observed at frame $t$, we utilize the query embedding $q^{t}_{i}\in\Re^{D}$ (\ie, object queries from the mask transformer decoder~\cite{yu2022k}).
The memory buffer encodes appearance feature $\hat{q}^t_{i}$ as follows:
\begin{equation}
\label{eqn:mb_appearance_feature}
    \hat{q}^t_{i} = \begin{cases} (1-\lambda)\hat{q}^{t-1}_{i} + \lambda q^{t}_{i}, & \text{\small if $i$ \textbf{both} in memory and frame $t$,} \\
    q^{t}_{i}, & \text{\small else if $i$ \textbf{only} in frame $t$,} \\
    \hat{q}^{t-1}_{i}, & \text{\small else if $i$ \textbf{only} in memory,}
    \end{cases}
\end{equation}
where $\lambda$ is the moving average weight between the stored appearance feature in memory $\hat{q}^{t-1}_{i}$ and current appearance feature $q^{t}_{i}$. We set $\lambda$ to 0.8 as the default value.

Unlike other works~\cite{kim2020video,wu2022defense}, we additionally exploit the location feature of object $i$ observed at frame $t$, using its normalized bounding box (inferred from the predicted mask): $b^{t}_{i}$=$[x^{tl}_{i}/w, y^{tl}_{i}/h, x^{br}_{i}/w, y^{br}_{i}/h]$ $\in\Re^{4}$, where $(x^{tl}, y^{tl})$ and $(x^{br}, y^{br})$ are the x-y coordinates of top-left and bottom-right corners, and $w$ and $h$ denote the bounding box width and height, respectively.
The memory buffer then encodes the location features as follows:
\begin{equation}
\label{eq:mb_location_feature}
    \hat{b}^{t}_{i} = \begin{cases} b^{t}_{i}, & \text{\small if $i$ in frame $t$,}  \\
    \hat{b}^{t-1}_{i} + (\hat{b}^{t-1}_{i}-\hat{b}^{t-2}_{i}), & \text{\small else if $i$ \textbf{only} in memory.}
    \end{cases}
\end{equation}
As shown in the equations, if an object is detected, the memory buffer will use its latest normalized bounding box information.
If the object $i$ is not detected but it is stored in the memory (\eg, due to occlusion), we will "predict" its current location by assuming the object's moving velocity is constant, \ie, its location is shifted by $(\hat{b}^{t-1}_{i}-\hat{b}^{t-2}_{i})$ from its previous stored location $\hat{b}^{t-1}_{i}$.

Finally, the memory buffer stores both the appearance and location features $(\hat{q}_i, \hat{b}_i)$ for all $M$ objects detected until the current frame. In practice, we adopt the memory refreshing strategy~\cite{wu2022defense}, where the old objects, whose last appeared frame is $\tau$ frame behind the current frame, are removed from the memory buffer.
We empirically choose the optimal value for $\tau$ in our experiments.

{\bf Decoding Phase}~\quad
To specialize the memory buffer approach in our framework, we initially conduct the Video Stitching (VS) for \textit{short-term} association between clips.
Afterwards, we associate the objects stored in memory with the currently detected but \textit{unmatched} objects, aiming for \textit{long-term} association.
This \textit{hierarchical} matching mechanism forms our proposed \methodnamefull (\methodname).
Specifically, we compute the similarity function $f(i, j)$ between the currently \textit{unmatched} object $i$ (after VS) and the \textit{encoded} object $j$ in the memory as follows:
\begin{equation}
    f(i, j) = e^{-\|b_{i} - \hat{b}_{j}\|^2/T} \cdot cos(q_{i}, \hat{q}_{j}).
    \label{eqn:sim}
\end{equation}
We compute the negative $L_2$ distance between two normalized bounding boxes, weighted by a temperature $T$ for scaling the values between location and appearance similarity.
The appearance similarity is measured by the $cosine$ distance.
Then, we obtain a similarity matrix $\mathbf{S} \in\Re^{M\times N}$ between $M$ objects in memory and $N$ detected objects in the current frame.
To find the association, we perform Hungarian matching~\cite{Kuhn1955NAVAL} on $\mathbf{S}$.
Additionally, to filter out false associations, we only consider the matching with similarity value larger than a confidence threshold $\alpha$.
The unmatched objects in current frame are considered as new objects.
The proposed \methodname is illustrated in \figref{fig:method_lamb}.

{\bf Discussion}~\quad
Our proposed \methodname is partially inspired by the success of IDOL~\cite{wu2022defense} in video instance segmentation, and memory buffer has been proven effective in several recent works~\cite{Yang19ICCV,zeng2022motr,cai2022memot}.
However, there are two critical issues, if one na\"ively applies their memory buffer approach to our framework (we name this method as \methodbaselinefull (\methodbaseline) for our baseline).
First, the location feature is not exploited, but only the appearance feature. In a dynamic scene, object location plays an important role. The appearance feature becomes less reliable if the target object has been occluded for a long time.
Second, the memory size $M$ keeps growing as time goes by.
Even though this issue is slightly alleviated by the memory refreshing strategy, it still results in a large matching space between the stored $M$ objects in the memory and the currently detected $N$ objects, which subsequently makes the one-to-one matching harder.
To overcome the issues, our \methodname proposes a novel formulation to incorporate the location features (\equref{eq:mb_location_feature} and \equref{eqn:sim}), and additionally augments the matching accuracy by  performing the Video Stitching (VS) in the beginning of decoding phase, which effectively further reduces the matching space and improves the matching accuracy, as demonstrated in our ablation studies.

\subsection{Online Video Panoptic Segmentation}
\label{sec:online}
The meta architecture of \modelname enables a general framework for both online and near-online VPS.
When processing a clip of length one, our model performs online VPS.
Specifically, the model is trained frame-by-frame and evaluated sequentially with 
the assistance of clip-$k$MaX's general formulation.
Unlike the near-online setting, we skip the Video Stitching, which becomes infeasible in the online framework.
Afterwards, we apply our \methodname without any further modification.



%% file: sections/4.experiments.tex
\section{Experimental Results}
\label{sec:results}
We conduct experiments on two \textit{long sequences} Video Panoptic Segmentation datasets: KITTI-STEP~\cite{weber2021step} and VIPSeg~\cite{miao2022large}. Furthermore, we evaluate our method on a Video Semantic Segmentation (VSS) dataset: VSPW~\cite{miao2021vspw}.

\subsection{Datasets}

{\bf KITTI-STEP~\cite{weber2021step}} is a Video Panoptic Segmentation (VPS) dataset that contains long video sequences with average track length 51 frames and maximum 643 frames, presenting a challenging scenario for long-term association.
It contains 19 semantic classes, similar to Cityscapes~\cite{cordts2016cityscapes}, while only two classes (`pedestrians' and `cars') come with tracking IDs.
We adopt the Segmentation and Tracking Quality (STQ) as a metric for evaluation.

{\bf VIPSeg~\cite{miao2022large}} is a new large-scale Video Panoptic Segmentation (VPS) benchmark providing in-the-wild real-world scenarios with 232 scenes and 124 classes, 
Among them, 58 classes are annotated with tracking IDs.
The average sequence length is 24 frames per video.
We adopt the STQ and VPQ~\cite{kim2020video} metric for evaluation. 

{\bf VSPW~\cite{miao2021vspw}} is a recent large-scale Video Semantic Segmentation (VSS) dataset with 124 semantic classes.
VSPW adopts mIoU as the  evaluation metric.





\begin{table}
    \begin{subtable}{\linewidth}
      \centering
        \input{tables/kitti_val}
        \caption{KITTI-STEP \textit{val} set.}
    \end{subtable} 
    \begin{subtable}{\linewidth}
      \centering
\input{tables/kitti_test}
        \caption{KITTI-STEP \textit{test} set. $\dagger$: ICCV 2021 challenge winning entry.}
    \end{subtable}%
\caption{[VPS] KITTI-STEP \textit{val} and \textit{test} set results.}
\label{tab:result_kitti_val_test}
\end{table}

\subsection{Implementation Details}
The proposed \modelname is a unified approach for online and near-online VPS.
For the near-online setting, we employ a clip length of two with one overlapping frame between clips.
For the online setting, we set clip length to one and remove the video stitching strategy in the pipeline.



We employ two common backbones for both online and near-online settings: ResNet50~\cite{he2016deep} and ConvNeXt-L~\cite{liu2022convnet}.
We also experiment with Axial-ResNet50-B1~\cite{wang2020axial} backbone to fairly compare with TubeFormer~\cite{kim2022tubeformer}.
If not specified, we default to use ResNet50 for ablation studies.
Our \modelname is built with the official code-base~\cite{weber2021deeplab2}.
Closely following the prior works~\cite{weber2021step,kim2022tubeformer}, both the near-online and online models employ a specific pre-training protocol for KITTI-STEP, VIPSeg and VSPW.
They all commonly require ImageNet~\cite{russakovsky2015imagenet} pretrained checkpoint. VIPSeg and VSPW further require pre-training models on COCO~\cite{lin2014microsoft}. For KITTI-STEP, Cityscapes~\cite{cordts2016cityscapes} is additionally adopted as a pre-training dataset since they share a similar driving scene and class category. We note that our best backbone ConvNeXt-L~\cite{liu2022convnet} on KITTI-STEP uses both COCO~\cite{lin2014microsoft} and Cityscapes~\cite{cordts2016cityscapes} for pre-training.
\subsection{Main Results}

\begin{table}
    \begin{subtable}{0.99\linewidth}
      \centering
        \input{tables/vip_seg_val}
        \caption{VIPSeg \textit{val} set.}
    \end{subtable} 
    \begin{subtable}{0.98\linewidth}
      \centering
        \input{tables/vip_seg_test}
        \caption{VIPSeg \textit{test} set in the latest test server.}
    \end{subtable}%
\caption{[VPS] VIPSeg \textit{val} and \textit{test} set results.}
\label{tab:result_vipseg_val_test}
\end{table}

\begin{table}
    \begin{subtable}{\linewidth}
      \centering
        \input{tables/vspw_val}
        \caption{VSPW \textit{val} set.}
    \end{subtable} 
    \begin{subtable}{\linewidth}
      \centering
\input{tables/vspw_test}
        \caption{VSPW \textit{test} set.}
    \end{subtable}%
\caption{[VSS] VSPW \textit{val} and \textit{test} set results.}
\label{tab:result_vspw_val_Test}
\end{table}


{\bf [VPS] KITTI-STEP}~\quad
\tabref{tab:result_kitti_val_test} summarizes our performance on the KITTI-STEP \textit{val} and \text{test} sets.
On the validation set (\tabref{tab:result_kitti_val_test}~(a)), we compare methods in the two categories: online and near-online methods.
In the online setting, when using the standard ResNet50~\cite{he2016deep}, our \modelname (online) outperforms Video K-Net~\cite{li2022video} by \textbf{+2.5\%} STQ. To further push the envelope, our model, equipped with the modern backbone ConvNeXt-L~\cite{liu2022convnet}, achieves the new state-of-the-art with 76.5\% STQ.
In the near-online setting, when using ResNet50, our \modelname (near-online) significantly surpasses Motion-DeepLab~\cite{weber2021step} by \textbf{+16.2\%} STQ.
When employing Axial-ResNet50-B1~\cite{wang2020axial} backbone, \modelname (near-online) also outperforms TubeFormer~\cite{kim2022tubeformer} by \textbf{+2.8\%} STQ.
Finally, \modelname (near-online) with ConvNeXt-L further sets a new state-of-the-art performance with 78.9\% STQ, significantly outperforming current best result (TubeFormer with Axial-ResNet50-B3) by {\bf +4.6\%} STQ.
We observe the same trend on the test set (\tabref{tab:result_kitti_val_test}~(b)), where our model reaches 68.5\% STQ, significantly outperforming the prior arts TubeFormer~\cite{kim2022tubeformer}, Video K-Net~\cite{li2022video}, and Motion-DeepLab~\cite{weber2021step} by {\bf +3.2}\%, {\bf +5.5}\%, and {\bf +16.3}\% STQ, respectively.
Remarkably, our extremely simple model even outperforms the ICCV 2021 Challenge winning entry, UW\_IPL/ETRI\_AIRL~\cite{zhang2021u3dmolts} by {\bf +0.9\%} STQ, which exploits pseudo labels~\cite{zhu2019improving,chen2020naive} and adopts an exceedingly complicated system that not only consists of separate tracking, detection, and segmentation modules, but also requires 3D object and depth information.

{\bf [VPS] VIPSeg}~\quad
\tabref{tab:result_vipseg_val_test} (a) summarizes the results on the VIPSeg \textit{val} set.
In the online setting, our \modelname (online) with ResNet50 attains 38.7\% STQ / 36.8\% VPQ, significantly outperforming the prior art Video K-Net by {\bf +5.6\%} STQ / \textbf{+10.7\%} VPQ.
Using the ConvNeXt-L backbone, our model advances the new state-of-the-art to 49.4\% STQ / 49.4\% VPQ.
In the near-online setting,
when using ResNet50, our \modelname (near-online) surpasses  Clip-PanoFCN~\cite{miao2022large} by {\bf +8.4\%} STQ / {\bf +15.3\%} VPQ.
When using Axial-ResNet50-B1, \modelname (near-online) outperforms TubeFormer~\cite{kim2022tubeformer} by {\bf +6.0\%} STQ / {\bf +17.5\%} VPQ.
Our best setting with ConvNeXt-L backbone further advances the state-of-the-art to 51.7\% STQ / 51.9\% VPQ, outperforming TubeFormer with Axial-ResNet50-B3 by {\bf +10.2\%} STQ / {\bf +20.7\%} VPQ.
We also show the effectiveness of \modelname (near-online) on VIPSeg \textit{test} set in~\tabref{tab:result_vipseg_val_test} (b), where \modelname also sets a new state-of-the-art, outperforming TubeFormer~\cite{kim2022tubeformer} by \textbf{+8.5\%} STQ / \textbf{+18.2\%} VPQ. 


{\bf [VSS] VSPW}~\quad
\tabref{tab:result_vspw_val_Test} shows VSPW val and test set results. Our \modelname (online) outperforms TubeFormer~\cite{kim2022tubeformer} both in mIoU and VC metrics~\cite{miao2021vspw}.


\subsection{Ablation Studies}

{\bf Comparison with Normal Cross-Attention}~\quad
As discussed in~\secref{sec:nearonline_vps}, we deliberately design our clip-$k$MaX with the $k$-means cross-attention~\cite{yu2022k}, which we empirically found to be very effective for handling the extremely large sequence of spatially and temporally flattened clip features.
We now elaborate on the experiments and particularly compare with the normal (\ie, original) cross-attention~\cite{vaswani2017attention} as well as the advanced latent memory cross-attention~\cite{kim2022tubeformer} (\ie, the cross-attention mechanism used in TubeFormer~\cite{kim2022tubeformer}, which adopts latent memory to facilitate attention learning between video frames).

\tabref{tab:compare_cross} summarizes our findings.
To ensure the fairness, we employ the same backbone Axial-ResNet50-B1~\cite{wang2020axial} that has been pretrained on ImageNet-1K and Cityscapes.
The baseline, employing the normal cross-attention module, yields the performance of 68.4\% STQ.
The performance can be further improved by 1.6\% STQ, if we adopt the latent memory~\cite{kim2022tubeformer} in the cross-attention module.
By contrast, our clip-$k$MaX, adopting the $k$-means cross-attention mechanism, attains 73.9\% STQ, significantly outperforming the conventional cross-attention and latent memory cross-attention by {\bf +5.5\%} and {\bf +3.9\%} STQ, respectively.
The remarkable improvement is attributed to the effectiveness of $k$-means cross-attention that performs the cluster-wise argmax on cluster centers, while the normal cross-attention is performed \wrt the enormous long sequence of spatially and temporally flattened clip features, where each object query has difficulty in identifying the most distinguishable features among the abundant pixels.
Our results suggest that using $k$-means cross-attention can reduce the ambiguity in cross-attention between queries and large flattened clip features, resulting in a higher quality of video panoptic segmentation results. 
Additionally, we show that our proposed \methodname is complementary to clip-$k$MaX, which sets the best STQ performance (74.7\% STQ).

\input{tables/compare_w_cross}

{\bf Association Modules}~\quad
Our proposed \methodname exploits (1) Video Stitching (VS), (2) appearance feature, and (3) location feature, to perform the object association.
\begin{table}
    \begin{subtable}{.94\linewidth}
      \centering
        \input{tables/ablation_for_near_lamb_on_kitti}

        \caption{Near-online setting using clip-based trained models.}
    \end{subtable} 
    \begin{subtable}{.95\linewidth}
      \centering
        \input{tables/ablation_for_online_lamb_on_kitti}
        \caption{Online setting using image-based trained models.}
    \end{subtable}%
\caption{Ablation study on {\bf different association features}, including the Video Stitching strategy, appearance feature, and  location feature, on KITTI-STEP \textit{val} set. We note that employing different association features will only affect the association quality (AQ).
Our final \methodname setting is labeled with brown color, while video-stitching and na\"ive-MB baselines are
denoted in blue and red, respectively.
}

\label{tab:ablation_association_features}
\end{table}
In~\tabref{tab:ablation_association_features}, we carefully study the effect of each feature in \methodname under both near-online and online settings .
In the near-online setting (\tabref{tab:ablation_association_features}~(a)), when using these three features \textit{individually}, we discover that both VS and location feature are equally more effective than appearance feature.
We note that when using only the appearance, the method becomes the na\"ive-MB approach, used by other works~\cite{wu2022defense,huang2022minvis}.
Combining all of them leads to our best final setting, while taking out the location feature will degrade the AQ performance most.
This study demonstrates that our proposed location feature is the most effective feature among them.
In the online setting, since the VS strategy becomes infeasible, we only experiment with the appearance and location features.
As shown in~\tabref{tab:ablation_association_features}~(b), the pure image-based model, which does not exploit any association feature, attains the worst performance.
Interestingly, we notice that the appearance feature learned by the ResNet50~\cite{he2016deep} is less effective than ConvNeXt-L~\cite{liu2022convnet}.
When the appearance feature is less effective (\eg, when using ResNet50), it is better to just use the location feature for association.
On the other hand, when the appearance feature is sufficiently informative (\eg, when using ConvNeXt-L), the best performance is obtained by using both appearance and location features.

\input{tables/space_reduction}

\input{tables/hyperparameter_stability_short}

{\bf Analysis on Memory Matching Space}~\quad
As discussed in~\secref{sec:naive_mb}, we address the limitations of the previous memory buffer approach~\cite{wu2022defense}, referred as na\"ive-MB.
One of the limitations of na\"ive-MB is the huge matching space in memory decoding, which increases the difficulty of matching and thus results in low association quality.
From that perspective, we empirically prove that our hierarchical matching scheme, \methodname, can effectively reduce the matching space size as shown in~\tabref{tab:space_reduction}.
To do so, we calculate the size of the similarity matrix $\mathbf{S}$ (\ie, $M\times N$, where there are $M$ objects in the memory and $N$ detected objects in the current frame) to quantitatively measure the matching space size.
We note that modern approaches~\cite{wu2022defense} adopt a memory refreshing strategy, where the old objects stored in the memory will be removed if they are $\tau$-frame older than the current frame, which, to some degree, alleviates the issue of large matching space.
However, we will show that using the memory refreshing strategy alone is not sufficient to reduce the matching space size.
We compare the matching space between na\"ive-MB and our \methodname under two cases of $\tau$, which is the hyper-parameter to refresh out the old objects in the memory, affecting the matching space size.
In the first case, we set $\tau$ to 1000, which mimics the ideal scenario, where we have a very large memory and the old objects are barely removed, aiming to exclude the effect of refreshing strategy and focus on the memory buffer approach itself.
As shown in the table, we can observe that \methodname can greatly improve the matching space efficiency by a healthy margin (\ie, 3.4$\times$ smaller and 3.6$\times$ smaller in average and max values, respectively).
In the second case, $\tau$ is set to be the optimal value for each memory buffer approach (\ie, 1 for na\"ive-MB and 10 for \methodname).
As shown in the table, the memory refreshing strategy effectively reduces the matching space size for na\"ive-MB.
However, our \methodname still outperforms na\"ive-MB by achieving 9.1$\times$ and 8.2$\times$ more efficient matching space in average and max value, respectively.

{\bf Memory-related Hyper-parameters}~\quad
Our proposed memory module \methodname contains two hyper-parameters: $\tau$ (for refreshing old objects in the memory buffer) and $\alpha$ (confidence threshold for matching).
In~\tabref{tab:ablation_hyper_parameter}, we ablate their effects on our \methodname and the baseline na\"ive-MB.
As shown in the table, our \methodname not only performs better, but also is more robust to the hyper-parameter values than na\"ive-MB.
More concretely, when computing the mean and standard deviation (std) for the obtained AQ \wrt. different $\tau$ and $\alpha$, our \methodname achieves a mean of 73.4 and a std of 0.4, while the baseline na\"ive-MB attains a \textit{lower} mean of 61.5 and a \textit{higher} std of 5.8.
We think the robustness of \methodname could be attributed to its efficient hierarchical matching scheme, which avoids the ambiguity caused by the large matching space.

\input{tables/ablation_for_temperature}

\input{tables/ablation_for_lambda}

\begin{figure*}[t!]
\begin{center}
\includegraphics[width=0.88\linewidth]{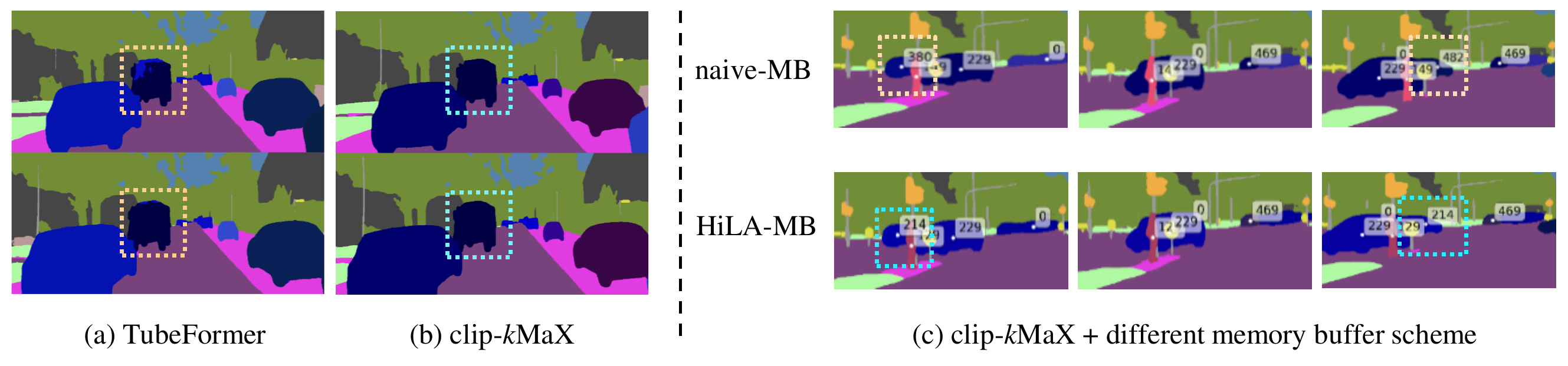}
\caption{
{\bf Visualization results} on KITTI-STEP \textit{val} set.
The proposed within-clip segmenter, clip-$k$MaX, segments objects in a clip better than the state-of-art TubeFormer ((a) \emph{vs}. (b)). In (c),
the proposed cross-clip associater, \methodname (\methodnamefull), associates occluded objects better than the baseline na\"ive-MB, which exploits only appearance features.
}
\label{fig:visualization}
\end{center}
\end{figure*}

\begin{figure*}[!th]
\begin{center}
\includegraphics[width=0.9\linewidth]{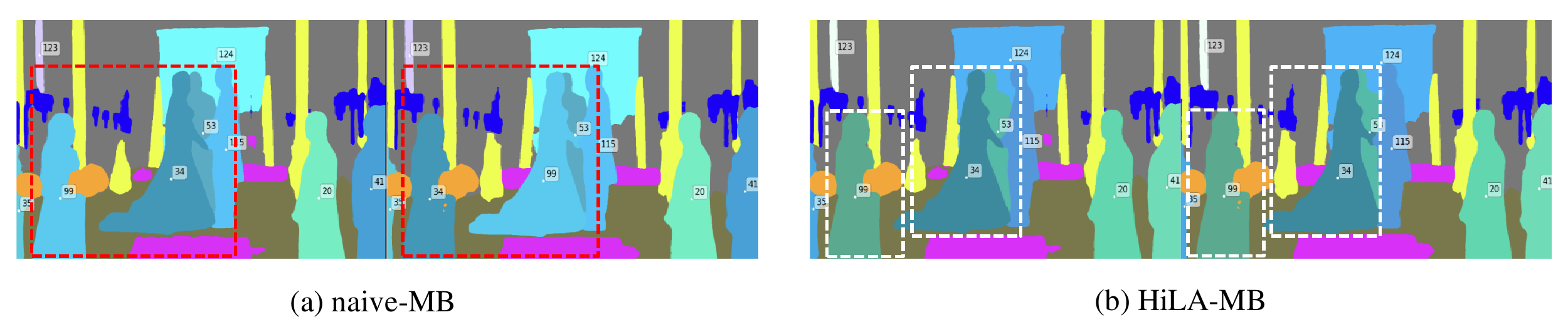}
\caption{
{\bf Visualization results} on VIPSeg \textit{val} set.
The baseline na\"ive-MB, only exploiting the appearance feature, fails to associate the same person, as neighboring people have similar appearance features.
On the other hand, our \methodname, exploiting both appearance and location features, successfully associates the same person.
}
\label{fig:vip_seg_vis}
\end{center}
\vspace{-1mm}
\end{figure*}

{\bf Feature-related Hyper-parameter}~\quad
We adopt a temperature $T$ to scale the values between the location and appearance features (see ~\equref{eqn:sim}).
As shown in~\tabref{tab:ablation_temperature}, our model is robust to the different values of $T$.
We thus default its value to 1 for simplicity.
Additionally, as shown in~\tabref{tab:ablation_lambda}, our model is also robust to the different values of $\lambda$, which balances the weight between the stored appearance feature in memory and the current one (see ~\equref{eqn:mb_appearance_feature}).

\subsection{Visualization Analysis}
{\bf Qualitative Results}~\quad
We visualize results in~\figref{fig:visualization} for KITTI-STEP.
clip-$k$MaX performs better than the state-of-the-art TubeFormer~\cite{kim2022tubeformer} for consistent segmentation between frames in a clip.
The proposed \methodname enables long-term association, successfully re-identifying the occluded car object (ID 214), while the baseline na\"ive-MB fails, since it only exploits the appearance feature.
Additionally, we show some visualization results in~\figref{fig:vip_seg_vis} for VIPSeg, where the baseline na\"ive-MB fails to associate persons in a crowd, since they have similar appearance features.
On the other hand, our \methodname correctly associates the same person by effectively exploiting both the appearance and location features. Finally, our \modelname (consisting of clip-$k$MaX and \methodname) demonstrates more clear and consistent
video results than the baselines as provided in \href{https://youtu.be/gK3bUCNnvGA}{https://youtu.be/gK3bUCNnvGA}.

\begin{figure*}[!th]
\begin{center}
\includegraphics[width=0.85\linewidth]{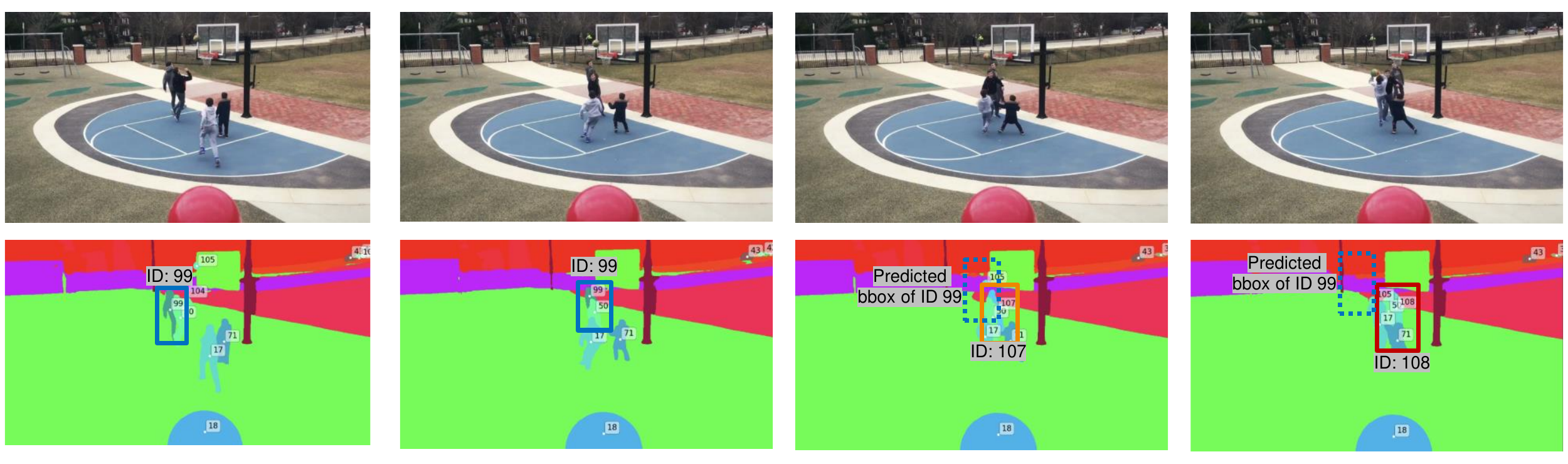}
\caption{
{\bf Failure case} on VIPSeg \textit{val} set.
The target object is initially assigned with ID 99.
Its ID switches to 107 and 108 in frame 3 and frame 4, respectively.
Our method fails to track the target object, because it is heavily occluded and moves at a large random pace, making both appearance and location features unreliable.
}
\label{fig:vip_seg_fail}
\end{center}
\vspace{-0mm}
\end{figure*}

\begin{figure}[t]
\begin{center}
\includegraphics[width=0.8\linewidth]{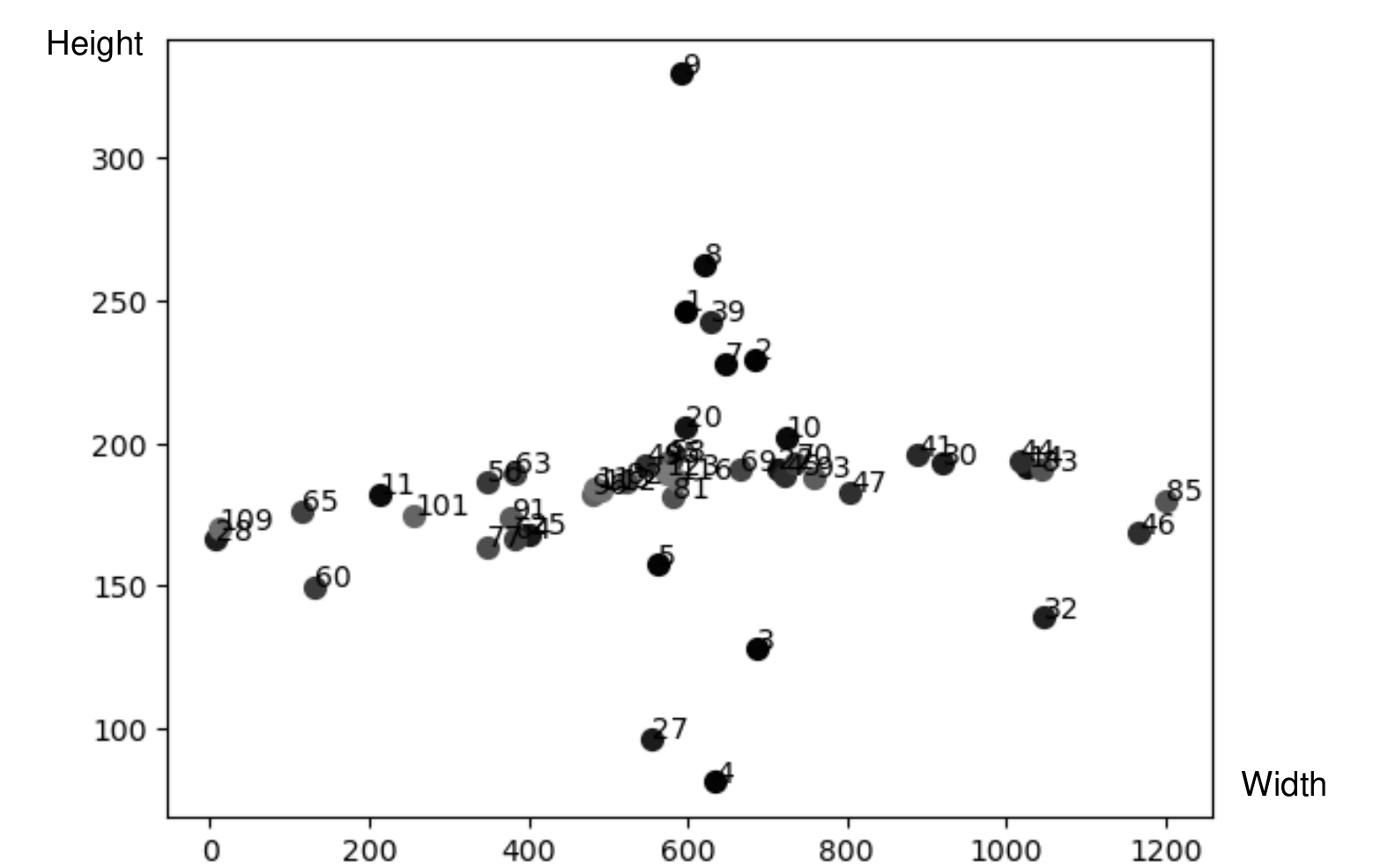}
\caption{
{\bf Query visualization} on KITTI-STEP \textit{val} set. We use \modelname with Axial-ResNet50-B1 backbone that is trained on KITTI-STEP and then plot the location of averaged mask center (including all stuff and things) predicted by each query.  
}
\label{fig:query_space}
\end{center}
\vspace{-0mm}
\end{figure}

{\bf Structural Prior Learned by Queries}~\quad
We observe that the object queries learned by our \modelname demonstrate a structural prior that a particular query will respond to objects around a specific location on the image plane.
To visualize the structural prior, for each query, we compute the mean location center of all its segmented objects in the whole KITTI-STEP validation set, and show the scatter plot in~\figref{fig:query_space}.
As shown in the figure, each object query is responsible to segment objects around a specific location on the image plane.
Interestingly, the object queries are scattered mostly along a vertical and a horizontal line, showing the property of ego-centric car in KITTI-STEP, where the street-view images are collected by a driving car.

{\bf Failure Case and Limitation}~\quad
We analyze the failure mode of our \modelname in~\figref{fig:vip_seg_fail}.
The first row and second row are video frames and corresponding video panoptic results with our \modelname, respectively. We  observe that a person initially assigned with ID number 99 until frame 2 is re-assigned with different ID numbers, \ie, 107 (in frame 3) and 108 (in frame 4).
The ID switch could be attributed to two reasons.
First, the appearance feature of the occluded person (\ie, person ID 107 in frame 3) is not reliable, as most of its discriminative appearance regions are occluded.
Second, the target object demonstrates a large random movement, violating our slow linear motion assumption encoded by the location feature.

%% file: tables/kitti_val.tex

\resizebox{0.91\linewidth}{!}{
\begin{tabular}{lc|cc|c}
method & backbone & SQ & AQ & STQ\\
\shline
\multicolumn{5}{l}{\setting{\textit{\textbf{online methods}}}} \\
Video K-Net~\cite{li2022video} & ResNet50 & 71.0 & 70.0 & 71.0 \\
Video K-Net~\cite{li2022video} & Swin-L & 75.0 & 73.0 & 74.0 \\
\hline
\modelname (online)   & ResNet50 & 75.0 & 72.0  & 73.5 \\
\modelname (online)  & ConvNeXt-L & \textbf{77.2} & \textbf{75.7} & \textbf{76.5} \\
\hline \hline
\multicolumn{5}{l}{\setting{\textit{\textbf{near-online methods}}}} \\
Motion-DeepLab~\cite{weber2021step} & ResNet50 & 67.0 & 51.0 & 58.0\\
TubeFormer~\cite{kim2022tubeformer} & Axial-ResNet50-B1 & 78.1 & 68.6 & 73.2 \\
TubeFormer~\cite{kim2022tubeformer} & Axial-ResNet50-B3 & 78.3 & 70.0 & 74.3 \\
\hline
\modelname (near-online)   & ResNet50  & 74.2  & 74.2  & 74.2 \\
\modelname (near-online)   & Axial-ResNet50-B1  & 75.8 & 76.3 & 76.0 \\
\modelname (near-online)    & ConvNeXt-L  & \textbf{79.0} & \textbf{78.8} & \textbf{78.9} \\
\end{tabular}}

%% file: tables/kitti_test.tex
\resizebox{0.72\linewidth}{!}{
\begin{tabular}{l|cc|c}
method & SQ & AQ & STQ\\
\shline
Motion-DeepLab~\cite{weber2021step} & 59.8 & 45.6 & 52.2\\
Video K-Net~\cite{li2022video} & 65.0 & 60.0 & 63.0\\
TubeFormer~\cite{kim2022tubeformer} & {\bf 70.3} & 60.6 & 65.3 \\
UW\_IPL/ETRI\_AIRL~\cite{zhang2021u3dmolts}$\dagger$
                                    &  64.0     & {\bf 71.3}  & 67.6     \\
\hline
\hline
\modelname (near-online)    & 69.8  & 67.2 & {\bf 68.5}  \\
\end{tabular}}

%% file: tables/vip_seg_val.tex

\resizebox{0.94\linewidth}{!}{
\begin{tabular}{lc|cc|c ||c}
method & backbone & SQ & AQ & STQ & VPQ\\
\shline
\multicolumn{6}{l}{\setting{\textit{\textbf{online methods}}}} \\
VPSNet-FuseTrack~\cite{kim2020video} & ResNet50 & - & - & 20.8 & 17.0 \\
VPSNet-SiamTrack~\cite{woo2021learning} & ResNet50 & - & - & 21.1 & 17.2\\
Video K-Net~\cite{li2022video} (arXiv version)  & ResNet50      & - & - & 33.1 & 26.1\\ 
Video K-Net~\cite{li2022video} (arXiv version)  & Swin-base      & - & - & 46.3 & 39.8\\ 
\hline 
\modelname (online)    & ResNet50 & 46.3 & 32.4 & 38.7 & 36.8\\
\modelname (online)    & ConvNeXt-L & \textbf{60.7} & \textbf{40.2} & \textbf{49.4} & \textbf{49.4} \\
\hline \hline
\multicolumn{6}{l}{\setting{\textit{\textbf{near-online methods}}}} \\
ViP-DeepLab~\cite{vip_deeplab} & ResNet50 & - & - & 22.0 & 16.0\\
Clip-PanoFCN~\cite{miao2022large}  & ResNet50      & - & - & 31.5 & 22.9 \\
TubeFormer~\cite{kim2022tubeformer} (arXiv version)  & Axial-ResNet50-B1      & 50.3 & 31.6 & 39.8 & 29.2  \\
TubeFormer~\cite{kim2022tubeformer} (arXiv version)  & Axial-ResNet50-B3      & 53.0 & 32.5  & 41.5 & 31.2 \\ 
\hline
\modelname (near-online) & ResNet50    & 45.1 & 35.3 & 39.9 & 38.2\\ 
\modelname (near-online) & Axial-ResNet50-B1    & 55.6 & 37.8 & 45.8 & 46.7 \\
\modelname (near-online) & ConvNeXt-L    & \textbf{61.4} & \textbf{43.5} & \textbf{51.7} & \textbf{51.9} \\
\end{tabular}}

%% file: tables/vip_seg_test.tex


\resizebox{0.78\linewidth}{!}{
\begin{tabular}{l|cc}
method & STQ & VPQ\\
\shline
Clip-PanoFCN~\cite{miao2022large} & 25.0 & 22.9  \\
TubeFormer~\cite{kim2022tubeformer} (arXiv version) & 38.6 & 26.8  \\
\hline
\hline
\modelname (near-online)    &  \textbf{47.1} & \textbf{45.0}     \\
\end{tabular}}

%% file: tables/vspw_val.tex


\resizebox{0.8\linewidth}{!}{
\begin{tabular}{lc|ccc}
method               &backbone          & mIoU      & VC8       & VC16       \\
\shline
TCB~\cite{miao2021vspw} & ResNet101   & 37.8     & 87.9     & 84.0     \\

TubeFormer~\cite{kim2022tubeformer}                  &Axial-ResNet50-B1 & 58.0 & 90.1 & 86.8 \\
TubeFormer~\cite{kim2022tubeformer}                  &Axial-ResNet50-B4 & 63.2 & \textbf{92.1 } & 88.0 \\
\hline
\hline
\modelname (online) & ResNet50 & 44.3  & 86.0  & 81.4 \\
\modelname (online) & Axial-ResNet50-B1 & 59.8  & 89.2  & 85.6  \\
\modelname (online) & ConvNeXt-L & \textbf{63.6}  & 91.8  & \textbf{88.6}
\end{tabular}}

%% file: tables/vspw_test.tex
\resizebox{0.81\linewidth}{!}{
\begin{tabular}{l|ccc}
method & mIoU & VC8 & VC16\\
\shline
TCB~\cite{miao2021vspw} & 32.6 & 79.5 & 73.2  \\
TubeFormer~\cite{kim2022tubeformer} (arxiv version) & 53.0 & 90.2 & 86.4   \\
\hline
\hline
\modelname (online)  & \textbf{54.9}  & \textbf{91.6} & \textbf{88.6}     \\
\end{tabular}}

%% file: tables/compare_w_cross.tex
\begin{table}[t!]
\centering
\resizebox{1.03\linewidth}{!}{
\begin{tabular}{c|l|c}
\multicolumn{1}{l|}{backbone}      & method                                 & STQ \\ \shline
\multirow{5}{*}{Axial-ResNet50-B1} & normal cross-attention (baseline)               & 68.4        \\ \cline{2-3} 
                                   & latent memory cross-attention             & 70.0     \\
 \cline{2-3} 
                                   & \textbf{$k$-means cross-attention (clip-$k$MaX) }   & 73.9     \\
                                   & \textbf{+ \methodname} &  74.7   
\end{tabular}
}
\caption{
\textbf{Comparison with normal cross-attention.} The $k$-means cross-attention adopted by our proposed clip-$k$MaX achieves the best STQ than the normal cross-attention and latent memory cross-attention, demonstrating the effectiveness of $k$-means cross-attention in video understanding task.
}
\label{tab:compare_cross} 
\end{table}

%% file: tables/ablation_for_near_lamb_on_kitti.tex
\resizebox{0.85\linewidth}{!}{
\begin{tabular}{l|ccc|c}
               & \multicolumn{3}{c|}{association features}    & \\
method         & video-stitching    & appearance      & location        & AQ       \\
\hline

        & \vsbaseline{\checkmark}    & \vsbaseline{}   & \vsbaseline{}             & \vsbaseline{72.3}        \\
        & \idolbaseline{}    & \idolbaseline{\checkmark}  & \idolbaseline{}                & \idolbaseline{71.4} \\
        &     &      & \checkmark       &  72.3      \\
\modelname  &     & \checkmark & \checkmark           &  73.8       \\
(near-online)        & \checkmark    & \checkmark &        &    72.1    \\
        &  \checkmark   &  & \checkmark  &   73.6     \\
        &  \baseline{\checkmark}    &  \baseline{\checkmark} & \baseline{\checkmark}     &  \baseline{\textbf{74.2}} \\
\end{tabular}}

%% file: tables/ablation_for_online_lamb_on_kitti.tex
\resizebox{0.82\linewidth}{!}{
\begin{tabular}{l|c|cc|c}
                &   & \multicolumn{2}{c|}{association features}  & \\
method & backbone  & appearance      & location        & AQ      \\
\hline

    & \multirow{4}{*}{ResNet50}   &     &                     & 10.0       \\
    &   & \idolbaseline{\checkmark}    & \idolbaseline{}           & \idolbaseline{33.8}    \\
    &    & \baseline{}            & \baseline{\checkmark}       &   \baseline{\textbf{72.0}}   \\
\modelname    &    & \checkmark    & \checkmark       &   66.4   \\
    \cline{2-5}\cline{2-5}
(online)    &  \multirow{4}{*}{ConvNeXt-L}    &     &                 & 10.4      \\
    &   & \idolbaseline{\checkmark}    & \idolbaseline{} & \idolbaseline{61.6}      \\
    &    &               & \checkmark          &   74.0   \\
    &    & \baseline{\checkmark}    & \baseline{\checkmark}       &    \baseline{\textbf{75.7}}      \\
\end{tabular}}

%% file: tables/space_reduction.tex
\begin{table}[t!]
\centering
\resizebox{0.8\linewidth}{!}{
\begin{tabular}{c|llll}
\multirow{5}{*}{memory}       & \multicolumn{4}{c}{size of matching space $\mathbf{S}$ (\ie, $M\times N$)}                                                     \\ \cline{2-5} 
                              & \multicolumn{2}{c|}{$\tau=1000$}                                & \multicolumn{2}{c}{\begin{tabular}[c]{@{}c@{}}$\tau=best$\\ (na\"ive-MB: $\tau=1$, \\ \methodname: $\tau=10$)\end{tabular}}           \\ \cline{2-5} 
                              & \multicolumn{1}{c}{average} & \multicolumn{1}{c}{max} & \multicolumn{1}{c}{average} & \multicolumn{1}{c}{max} \\ \hline
\multicolumn{1}{l|}{na\"ive-MB} & \multicolumn{1}{c|}{67.1}        & \multicolumn{1}{c|}{336}    & \multicolumn{1}{c|}{25.1}        & \multicolumn{1}{c}{196}     \\ \hline
\multicolumn{1}{l|}{\methodname}    & \multicolumn{1}{c|}{19.7}        & \multicolumn{1}{c|}{94}    & \multicolumn{1}{c|}{2.8}        &  \multicolumn{1}{c}{24}   
\end{tabular}}
\caption{\textbf{Quantitative analysis on matching space size} between na\"ive-MB and our \methodname.
The size of matching space $\mathbf{S}$ could help us understand the difficulty of matching $M$ objects in the memory with the detected $N$ objects in the current frame.
$\tau$ is the hyper-parameter to refresh out the old objects.
We consider two cases, where $\tau=1000$ to mimic the case where we barely remove the old objects, and $\tau=best$ uses the best hyper-parameter value for each setting.
}
\label{tab:space_reduction}
\vspace{-5mm}
\end{table}

%% file: tables/hyperparameter_stability_short.tex
\begin{table*}[ht]
\centering
\begin{tabular}{ccccc|c}
\resizebox{0.55\linewidth}{!}{
\begin{tabular}{lcccc|c}
\multicolumn{1}{l}{} & \multicolumn{4}{c|}{hyper-parameter set [$\tau$ / $\alpha$]} & row-wise \\ \cline{2-5}
AQ (\%) & \cellcolor[HTML]{e2edf7}[1 / 0.6] & \cellcolor[HTML]{e2edf7} [3 / 0.6] & \cellcolor[HTML]{e2edf7}[10 / 0.6] & \cellcolor[HTML]{e2edf7}[20 / 0.6] & (mean / std)                     \\ \cline{2-5}
\methodbaseline & 69.5  & 67.1  & 54.3  &  47.4 &  59.6 / 10.5                     \\
\methodname & 72.7 & 73.0   & 73.9 & 73.7  &    73.3 / 0.6                     \\\cline{2-5}
& \cellcolor[HTML]{e2edf7}[1 / 0.7] & \cellcolor[HTML]{e2edf7} [3 / 0.7] & \cellcolor[HTML]{e2edf7}[10 / 0.7] & \cellcolor[HTML]{e2edf7}[20 / 0.7] &                                  \\ \cline{2-5}
\methodbaseline & 69.9 & 68.8  & 57.4 & 50.7 & 61.7 / 9.3                       \\
\methodname & 72.8  & 73.4   & \baseline{74.2} & 74.2  & 73.6 / 0.7                        \\\cline{2-5}
& \cellcolor[HTML]{e2edf7}[1 / 0.8] & \cellcolor[HTML]{e2edf7}[3 / 0.8] & \cellcolor[HTML]{e2edf7}[10 / 0.8] & \cellcolor[HTML]{e2edf7}[20 / 0.8] & \\ \cline{2-5}
\methodbaseline & \idolbaseline{71.4} & 71.3  & 64.9  & 59.5  &    66.8 / 5.7                   \\
\multicolumn{1}{l}{\methodname}                                                       &    72.6                                  &       73.2                                & 73.6 & 73.6 & 73.3 / 0.5 \\ \hline
                                                                                     & \multicolumn{1}{c}{69.7 / 1.0}          & \multicolumn{1}{l}{68.0 / 2.1}          & \multicolumn{1}{l}{55.8 / 5.5}          & \multicolumn{1}{c|}{49.1 / 6.3}          & \multicolumn{1}{c}{61.5 / 5.8} \\
\multirow{-2}{*}{\begin{tabular}[c]{@{}c@{}}column-wise\\ (mean / std)\end{tabular}} & \multicolumn{1}{c}{72.7 / 0.1}          & \multicolumn{1}{c}{73.2 / 0.2}          & \multicolumn{1}{c}{74.1 / 0.3}          & \multicolumn{1}{c|}{74.0 / 0.3}          & \multicolumn{1}{c}{73.4 / 0.4}
\end{tabular}
}
\input{figs/hyperparameter_graph}
\end{tabular}
\caption{Ablation study on \textbf{stability of \modelname using different memory-related hyper-parameter sets ($\tau$ for memory-refreshing and $\alpha$ for confidence threshold)} on KITTI-STEP \textit{val} set.
We vary $\tau \in \{1, 3, 10, 20\}$ (different columns in the table) and $\alpha \in \{0.6, 0.7, 0.8\}$ (different rows in the table).
We compute the mean and standard deviation column-wise (fixed $\tau$ and varied $\alpha$), row-wise (varied $\tau$ and fixed $\alpha$), and table-wise (varied $\tau$ and $\alpha$).
We plot the mean and standard deviation for the whole table on the right.
The proposed \methodname is more robust to the hyper-parameter values than the na\"ive-MB approach.
Our final \methodname setting and the na\"ive-MB baseline are labeled with brown and red color, respectively.
} 
\label{tab:ablation_hyper_parameter}
\vspace{-2mm}
\end{table*}

%% file: figs/hyperparameter_graph.tex
\pgfplotsset{compat=1.7}


\begin{tikzpicture}[baseline=(current axis.outer west)]
\begin{axis}
[
width=0.28\textwidth, 
legend pos=outer north east,
enlargelimits={abs=1.0},
ybar=0pt,
bar width=0.5,
xtick={0.5,1.5,...,3.5},
xticklabels={VS, \methodbaseline, \methodname},
x tick label as interval,
grid=both,
grid style={line width=.1pt, draw=gray!20},
]

\addplot+[error bars/.cd,
y dir=both,y explicit]
coordinates {
    (1.5, 72.3) +- (0.0, 0.0)};
\addplot+[error bars/.cd,
y dir=both,y explicit]
coordinates {
    (2.0, 61.52) +- (0.0, 5.76)};
\addplot+[error bars/.cd,
y dir=both,y explicit]
coordinates {
    (2.5,73.37) +- (0.0, 0.39)};
\end{axis}
\end{tikzpicture}

%% file: tables/ablation_for_temperature.tex
\begin{table}[t!]
\centering
\resizebox{0.8\linewidth}{!}{
\begin{tabular}{c|c|c|c}
method  & $ T $ & AQ & STQ\\
\thickhline
\multirow{3}{*}{\modelname (near-online)}  & 0.5 & 73.95  & 74.10  \\
  & \baseline{1.0} & \baseline{74.22} & \baseline{74.23}  \\
  & 1.5 & 74.30 & 74.27 \\
\end{tabular}}
\caption{Ablation study on \textbf{temperature $T$}, which scales the values between location and appearance features. Our final setting is labeled with brown color.
In this table, we show results up to two decimal points to more clearly see the robustness to $T$.
}
\label{tab:ablation_temperature}
\end{table}

%% file: tables/ablation_for_lambda.tex
\begin{table}[t!]
\centering
\resizebox{0.8\linewidth}{!}{
\begin{tabular}{c|c|c|c}
method  & $\lambda$ & AQ & STQ\\
\thickhline
\multirow{6}{*}{\modelname (near-online)}  & 0.0 & 73.34  & 73.80 \\
  & 0.5 & 74.19  & 74.21  \\
  & 0.7 & 74.22 & 74.23  \\
  & \baseline{0.8} & \baseline{74.22} & \baseline{74.23}  \\
  & 0.9 & 74.22 & 74.23  \\
  & 1.0 & 74.18 & 74.21   \\
\end{tabular}}
\caption{Additional analysis on \textbf{moving average weight $\lambda$}, which balances the stored appearance feature in the memory and current appearance feature.
Our final setting is labeled with brown color.
In this table, we show results up to two decimal points to more clearly see the robustness to $\lambda$.
}
\label{tab:ablation_lambda}
\end{table}

%% file: sections/5.conclusion.tex
\section{Conclusion}
\label{sec:conclusion}
\vspace{-1.1mm}
In this work, we proposed \modelname, a unified framework for online and near-online Video Panoptic Segmentation (VPS) model with two modules: clip-$k$MaX and \methodname.
The clip-$k$MaX utilizes object queries as cluster centers to group pixels of the same object within a clip, while the \methodname is a novel and robust memory module for both short- and long-term association with a hierarchical matching scheme. 
The effectiveness of our approach is demonstrated on the KITTI-STEP, VIPSeg and VSPW datasets.
We hope our study will inspire more future research on a unified framework for online and near-online VPS.

%% file: arxiv.bbl
\begin{thebibliography}{10}\itemsep=-1pt

\bibitem{bertasius2020classifying}
Gedas Bertasius and Lorenzo Torresani.
\newblock Classifying, segmenting, and tracking object instances in video with
  mask propagation.
\newblock In {\em Proceedings of the IEEE/CVF Conference on Computer Vision and
  Pattern Recognition}, 2020.

\bibitem{cai2022memot}
Jiarui Cai, Mingze Xu, Wei Li, Yuanjun Xiong, Wei Xia, Zhuowen Tu, and Stefano
  Soatto.
\newblock Memot: Multi-object tracking with memory.
\newblock In {\em Proceedings of the IEEE/CVF Conference on Computer Vision and
  Pattern Recognition}, pages 8090--8100, 2022.

\bibitem{carion2020end}
Nicolas Carion, Francisco Massa, Gabriel Synnaeve, Nicolas Usunier, Alexander
  Kirillov, and Sergey Zagoruyko.
\newblock End-to-end object detection with transformers.
\newblock In {\em Proceedings of the European Conference on Computer Vision},
  pages 213--229. Springer, 2020.

\bibitem{chen2020naive}
Liang-Chieh Chen, Raphael~Gontijo Lopes, Bowen Cheng, Maxwell~D Collins, Ekin~D
  Cubuk, Barret Zoph, Hartwig Adam, and Jonathon Shlens.
\newblock {Naive-Student: Leveraging Semi-Supervised Learning in Video
  Sequences for Urban Scene Segmentation}.
\newblock In {\em Proceedings of the European Conference on Computer Vision},
  2020.

\bibitem{chen2017deeplab}
Liang-Chieh Chen, George Papandreou, Iasonas Kokkinos, Kevin Murphy, and Alan~L
  Yuille.
\newblock Deeplab: Semantic image segmentation with deep convolutional nets,
  atrous convolution, and fully connected crfs.
\newblock {\em IEEE Transactions on Pattern Analysis and Machine Intelligence},
  40(4):834--848, 2017.

\bibitem{chen2018encoder}
Liang-Chieh Chen, Yukun Zhu, George Papandreou, Florian Schroff, and Hartwig
  Adam.
\newblock Encoder-decoder with atrous separable convolution for semantic image
  segmentation.
\newblock In {\em Proceedings of the European Conference on Computer Vision},
  pages 801--818, 2018.

\bibitem{cheng2021mask2former}
Bowen Cheng, Anwesa Choudhuri, Ishan Misra, Alexander Kirillov, Rohit Girdhar,
  and Alexander~G Schwing.
\newblock Mask2former for video instance segmentation.
\newblock {\em arXiv:2112.10764}, 2021.

\bibitem{cheng2019panopticworkshop}
Bowen Cheng, Maxwell~D Collins, Yukun Zhu, Ting Liu, Thomas~S Huang, Hartwig
  Adam, and Liang-Chieh Chen.
\newblock {Panoptic-DeepLab}.
\newblock In {\em ICCV COCO + Mapillary Joint Recognition Challenge Workshop},
  2019.

\bibitem{cheng2020panoptic}
Bowen Cheng, Maxwell~D Collins, Yukun Zhu, Ting Liu, Thomas~S Huang, Hartwig
  Adam, and Liang-Chieh Chen.
\newblock {Panoptic-DeepLab}: A simple, strong, and fast baseline for bottom-up
  panoptic segmentation.
\newblock In {\em Proceedings of the IEEE/CVF Conference on Computer Vision and
  Pattern Recognition}, 2020.

\bibitem{cordts2016cityscapes}
Marius Cordts, Mohamed Omran, Sebastian Ramos, Timo Rehfeld, Markus Enzweiler,
  Rodrigo Benenson, Uwe Franke, Stefan Roth, and Bernt Schiele.
\newblock The cityscapes dataset for semantic urban scene understanding.
\newblock In {\em Proceedings of the IEEE Conference on Computer Vision and
  Pattern Recognition}, pages 3213--3223, 2016.

\bibitem{dai2017deformable}
Jifeng Dai, Haozhi Qi, Yuwen Xiong, Yi Li, Guodong Zhang, Han Hu, and Yichen
  Wei.
\newblock Deformable convolutional networks.
\newblock In {\em Proceedings of IEEE International Conference on Computer
  Vision}, 2017.

\bibitem{gadde2017semantic}
Raghudeep Gadde, Varun Jampani, and Peter~V Gehler.
\newblock Semantic video {CNNs} through representation warping.
\newblock In {\em Proceedings of IEEE International Conference on Computer
  Vision}, 2017.

\bibitem{he2017mask}
Kaiming He, Georgia Gkioxari, Piotr Doll{\'a}r, and Ross Girshick.
\newblock Mask r-cnn.
\newblock In {\em Proceedings of IEEE International Conference on Computer
  Vision}, 2017.

\bibitem{he2016deep}
Kaiming He, Xiangyu Zhang, Shaoqing Ren, and Jian Sun.
\newblock Deep residual learning for image recognition.
\newblock In {\em Proceedings of the IEEE Conference on Computer Vision and
  Pattern Recognition}, pages 770--778, 2016.

\bibitem{heo2022generalized}
Miran Heo, Sukjun Hwang, Jeongseok Hyun, Hanjung Kim, Seoung~Wug Oh, Joon-Young
  Lee, and Seon~Joo Kim.
\newblock A generalized framework for video instance segmentation.
\newblock {\em arXiv preprint arXiv:2211.08834}, 2022.

\bibitem{heo2022vita}
Miran Heo, Sukjun Hwang, Seoung~Wug Oh, Joon-Young Lee, and Seon~Joo Kim.
\newblock Vita: Video instance segmentation via object token association.
\newblock {\em arXiv:2206.04403}, 2022.

\bibitem{huang2022minvis}
De-An Huang, Zhiding Yu, and Anima Anandkumar.
\newblock Minvis: A minimal video instance segmentation framework without
  video-based training.
\newblock {\em Advances in Neural Information Processing Systems}, 2022.

\bibitem{hwang2021video}
Sukjun Hwang, Miran Heo, Seoung~Wug Oh, and Seon~Joo Kim.
\newblock Video instance segmentation using inter-frame communication
  transformers.
\newblock {\em Advances in Neural Information Processing Systems}, 2021.

\bibitem{jain2019accel}
Samvit Jain, Xin Wang, and Joseph~E Gonzalez.
\newblock Accel: A corrective fusion network for efficient semantic
  segmentation on video.
\newblock In {\em Proceedings of the IEEE/CVF Conference on Computer Vision and
  Pattern Recognition}, 2019.

\bibitem{kim2020video}
Dahun Kim, Sanghyun Woo, Joon-Young Lee, and In~So Kweon.
\newblock Video panoptic segmentation.
\newblock In {\em Proceedings of the IEEE/CVF Conference on Computer Vision and
  Pattern Recognition}, pages 9859--9868, 2020.

\bibitem{kim2022tubeformer}
Dahun Kim, Jun Xie, Huiyu Wang, Siyuan Qiao, Qihang Yu, Hong-Seok Kim, Hartwig
  Adam, In~So Kweon, and Liang-Chieh Chen.
\newblock Tubeformer-deeplab: Video mask transformer.
\newblock In {\em Proceedings of the IEEE/CVF Conference on Computer Vision and
  Pattern Recognition}, pages 13914--13924, 2022.

\bibitem{kirillov2019panoptic}
Alexander Kirillov, Ross Girshick, Kaiming He, and Piotr Doll{\'a}r.
\newblock Panoptic feature pyramid networks.
\newblock In {\em Proceedings of the IEEE/CVF Conference on Computer Vision and
  Pattern Recognition}, 2019.

\bibitem{kirillov2018panoptic}
Alexander Kirillov, Kaiming He, Ross Girshick, Carsten Rother, and Piotr
  Doll{\'a}r.
\newblock Panoptic segmentation.
\newblock In {\em Proceedings of the IEEE/CVF Conference on Computer Vision and
  Pattern Recognition}, 2019.

\bibitem{Kuhn1955NAVAL}
Harold~W Kuhn.
\newblock The hungarian method for the assignment problem.
\newblock {\em Naval research logistics quarterly}, 2(1-2):83--97, 1955.

\bibitem{li2022video}
Xiangtai Li, Wenwei Zhang, Jiangmiao Pang, Kai Chen, Guangliang Cheng, Yunhai
  Tong, and Chen~Change Loy.
\newblock Video k-net: A simple, strong, and unified baseline for video
  segmentation.
\newblock In {\em Proceedings of the IEEE/CVF Conference on Computer Vision and
  Pattern Recognition}, pages 18847--18857, 2022.

\bibitem{li2018attention}
Yanwei Li, Xinze Chen, Zheng Zhu, Lingxi Xie, Guan Huang, Dalong Du, and
  Xingang Wang.
\newblock Attention-guided unified network for panoptic segmentation.
\newblock In {\em Proceedings of the IEEE/CVF Conference on Computer Vision and
  Pattern Recognition}, 2019.

\bibitem{li2021fully}
Yanwei Li, Hengshuang Zhao, Xiaojuan Qi, Liwei Wang, Zeming Li, Jian Sun, and
  Jiaya Jia.
\newblock Fully convolutional networks for panoptic segmentation.
\newblock In {\em Proceedings of the IEEE/CVF Conference on Computer Vision and
  Pattern Recognition}, 2021.

\bibitem{lin2014microsoft}
Tsung-Yi Lin, Michael Maire, Serge Belongie, James Hays, Pietro Perona, Deva
  Ramanan, Piotr Doll{\'a}r, and C~Lawrence Zitnick.
\newblock Microsoft coco: Common objects in context.
\newblock In {\em Proceedings of the European Conference on Computer Vision},
  2014.

\bibitem{liu2022convnet}
Zhuang Liu, Hanzi Mao, Chao-Yuan Wu, Christoph Feichtenhofer, Trevor Darrell,
  and Saining Xie.
\newblock A convnet for the 2020s.
\newblock In {\em Proceedings of the IEEE/CVF Conference on Computer Vision and
  Pattern Recognition}, pages 11976--11986, 2022.

\bibitem{meinhardt2022trackformer}
Tim Meinhardt, Alexander Kirillov, Laura Leal-Taixe, and Christoph
  Feichtenhofer.
\newblock Trackformer: Multi-object tracking with transformers.
\newblock In {\em Proceedings of the IEEE/CVF Conference on Computer Vision and
  Pattern Recognition}, pages 8844--8854, 2022.

\bibitem{miao2022large}
Jiaxu Miao, Xiaohan Wang, Yu Wu, Wei Li, Xu Zhang, Yunchao Wei, and Yi Yang.
\newblock Large-scale video panoptic segmentation in the wild: A benchmark.
\newblock In {\em Proceedings of the {IEEE} Conference on Computer Vision and
  Pattern Recognition}, 2022.

\bibitem{miao2021vspw}
Jiaxu Miao, Yunchao Wei, Yu Wu, Chen Liang, Guangrui Li, and Yi Yang.
\newblock Vspw: A large-scale dataset for video scene parsing in the wild.
\newblock In {\em Proceedings of the IEEE/CVF Conference on Computer Vision and
  Pattern Recognition}, 2021.

\bibitem{vip_deeplab}
Siyuan Qiao, Yukun Zhu, Hartwig Adam, Alan Yuille, and Liang-Chieh Chen.
\newblock Vip-deeplab: Learning visual perception with depth-aware video
  panoptic segmentation.
\newblock In {\em Proceedings of the IEEE/CVF Conference on Computer Vision and
  Pattern Recognition}, 2021.

\bibitem{russakovsky2015imagenet}
Olga Russakovsky, Jia Deng, Hao Su, Jonathan Krause, Sanjeev Satheesh, Sean Ma,
  Zhiheng Huang, Andrej Karpathy, Aditya Khosla, Michael Bernstein, et~al.
\newblock Imagenet large scale visual recognition challenge.
\newblock {\em International journal of computer vision}, 115(3):211--252,
  2015.

\bibitem{sun2020transtrack}
Peize Sun, Jinkun Cao, Yi Jiang, Rufeng Zhang, Enze Xie, Zehuan Yuan, Changhu
  Wang, and Ping Luo.
\newblock Transtrack: Multiple object tracking with transformer.
\newblock {\em arXiv:2012.15460}, 2020.

\bibitem{tian2020conditional}
Zhi Tian, Chunhua Shen, and Hao Chen.
\newblock Conditional convolutions for instance segmentation.
\newblock In {\em Proceedings of the European Conference on Computer Vision},
  2020.

\bibitem{vaswani2017attention}
Ashish Vaswani, Noam Shazeer, Niki Parmar, Jakob Uszkoreit, Llion Jones,
  Aidan~N Gomez, {\L}ukasz Kaiser, and Illia Polosukhin.
\newblock Attention is all you need.
\newblock {\em Advances in Neural Information Processing Systems}, 30, 2017.

\bibitem{voigtlaender2019mots}
Paul Voigtlaender, Michael Krause, Aljosa Osep, Jonathon Luiten, Berin
  Balachandar~Gnana Sekar, Andreas Geiger, and Bastian Leibe.
\newblock Mots: Multi-object tracking and segmentation.
\newblock In {\em Proceedings of the IEEE/CVF Conference on Computer Vision and
  Pattern Recognition}, 2019.

\bibitem{wang2021max}
Huiyu Wang, Yukun Zhu, Hartwig Adam, Alan Yuille, and Liang-Chieh Chen.
\newblock Max-deeplab: End-to-end panoptic segmentation with mask transformers.
\newblock In {\em Proceedings of the IEEE/CVF Conference on Computer Vision and
  Pattern Recognition}, pages 5463--5474, 2021.

\bibitem{wang2020axial}
Huiyu Wang, Yukun Zhu, Bradley Green, Hartwig Adam, Alan Yuille, and
  Liang-Chieh Chen.
\newblock Axial-deeplab: Stand-alone axial-attention for panoptic segmentation.
\newblock In {\em Proceedings of the European Conference on Computer Vision},
  2020.

\bibitem{wang2020solov2}
Xinlong Wang, Rufeng Zhang, Tao Kong, Lei Li, and Chunhua Shen.
\newblock Solov2: Dynamic and fast instance segmentation.
\newblock {\em Advances in Neural information processing systems}, 2020.

\bibitem{weber2021deeplab2}
Mark Weber, Huiyu Wang, Siyuan Qiao, Jun Xie, Maxwell~D Collins, Yukun Zhu,
  Liangzhe Yuan, Dahun Kim, Qihang Yu, Daniel Cremers, et~al.
\newblock Deeplab2: A tensorflow library for deep labeling.
\newblock {\em arXiv preprint arXiv:2106.09748}, 2021.

\bibitem{weber2021step}
Mark Weber, Jun Xie, Maxwell Collins, Yukun Zhu, Paul Voigtlaender, Hartwig
  Adam, Bradley Green, Andreas Geiger, Bastian Leibe, Daniel Cremers, Aljosa
  Osep, Laura Leal-Taixe, and Liang-Chieh Chen.
\newblock Step: Segmenting and tracking every pixel.
\newblock {\em Neural Information Processing Systems (NeurIPS) Track on
  Datasets and Benchmarks}, 2021.

\bibitem{woo2021learning}
Sanghyun Woo, Dahun Kim, Joon-Young Lee, and In~So Kweon.
\newblock Learning to associate every segment for video panoptic segmentation.
\newblock In {\em Proceedings of the IEEE/CVF Conference on Computer Vision and
  Pattern Recognition}, pages 2705--2714, 2021.

\bibitem{wu2022seqformer}
Junfeng Wu, Yi Jiang, Song Bai, Wenqing Zhang, and Xiang Bai.
\newblock Seqformer: Sequential transformer for video instance segmentation.
\newblock In {\em Proceedings of the European Conference on Computer Vision},
  pages 553--569. Springer, 2022.

\bibitem{wu2022defense}
Junfeng Wu, Qihao Liu, Yi Jiang, Song Bai, Alan Yuille, and Xiang Bai.
\newblock In defense of online models for video instance segmentation.
\newblock In {\em Proceedings of the European Conference on Computer Vision},
  pages 588--605. Springer, 2022.

\bibitem{wu2022efficient}
Jialian Wu, Sudhir Yarram, Hui Liang, Tian Lan, Junsong Yuan, Jayan Eledath,
  and Gerard Medioni.
\newblock Efficient video instance segmentation via tracklet query and
  proposal.
\newblock In {\em Proceedings of the IEEE/CVF Conference on Computer Vision and
  Pattern Recognition}, pages 959--968, 2022.

\bibitem{xiong19upsnet}
Yuwen Xiong, Renjie Liao, Hengshuang Zhao, Rui Hu, Min Bai, and Raquel~Urtasun
  Ersin~Yumer.
\newblock Upsnet: A unified panoptic segmentation network.
\newblock In {\em Proceedings of the IEEE/CVF Conference on Computer Vision and
  Pattern Recognition}, 2019.

\bibitem{xiong2019upsnet}
Yuwen Xiong, Renjie Liao, Hengshuang Zhao, Rui Hu, Min Bai, Ersin Yumer, and
  Raquel Urtasun.
\newblock {UPSNet}: A unified panoptic segmentation network.
\newblock In {\em Proceedings of the IEEE/CVF Conference on Computer Vision and
  Pattern Recognition}, 2019.

\bibitem{UNINEXT}
Bin Yan, Yi Jiang, Jiannan Wu, Dong Wang, Zehuan Yuan, Ping Luo, and Huchuan
  Lu.
\newblock Universal instance perception as object discovery and retrieval.
\newblock In {\em CVPR}, 2023.

\bibitem{Yang19ICCV}
Linjie Yang, Yuchen Fan, and Ning Xu.
\newblock {Video Instance Segmentation}.
\newblock In {\em Proceedings of IEEE International Conference on Computer
  Vision}, 2019.

\bibitem{yang2019deeperlab}
Tien-Ju Yang, Maxwell~D Collins, Yukun Zhu, Jyh-Jing Hwang, Ting Liu, Xiao
  Zhang, Vivienne Sze, George Papandreou, and Liang-Chieh Chen.
\newblock {DeeperLab}: Single-shot image parser.
\newblock {\em arXiv:1902.05093}, 2019.

\bibitem{yu2022cmt}
Qihang Yu, Huiyu Wang, Dahun Kim, Siyuan Qiao, Maxwell Collins, Yukun Zhu,
  Hartwig Adam, Alan Yuille, and Liang-Chieh Chen.
\newblock Cmt-deeplab: Clustering mask transformers for panoptic segmentation.
\newblock In {\em Proceedings of the IEEE/CVF Conference on Computer Vision and
  Pattern Recognition}, 2022.

\bibitem{yu2022k}
Qihang Yu, Huiyu Wang, Siyuan Qiao, Maxwell Collins, Yukun Zhu, Hartwig Adam,
  Alan Yuille, and Liang-Chieh Chen.
\newblock {k-means Mask Transformer}.
\newblock In {\em Proceedings of the European Conference on Computer Vision},
  pages 288--307. Springer, 2022.

\bibitem{zeng2022motr}
Fangao Zeng, Bin Dong, Yuang Zhang, Tiancai Wang, Xiangyu Zhang, and Yichen
  Wei.
\newblock Motr: End-to-end multiple-object tracking with transformer.
\newblock In {\em Proceedings of the European Conference on Computer Vision},
  pages 659--675. Springer, 2022.

\bibitem{zhang2021u3dmolts}
Haotian Zhang, Yizhou Wang, Zhongyu Jiang, Cheng-Yen Yang, Jie Mei, Jiarui Cai,
  Jenq-Neng Hwang, Kwang-Ju Kim, and Pyong-Kun Kim.
\newblock {U3D-MOLTS: Unified 3D Monocular Object Localization, Tracking and
  Segmentation}.
\newblock In {\em ICCV Segmenting and Tracking Every Point and Pixel: 6th
  Workshop on Benchmarking Multi-Target Tracking}, 2021.

\bibitem{zhang2021k}
Wenwei Zhang, Jiangmiao Pang, Kai Chen, and Chen~Change Loy.
\newblock K-net: Towards unified image segmentation.
\newblock {\em Advances in Neural Information Processing Systems},
  34:10326--10338, 2021.

\bibitem{zhao2022tracking}
Zelin Zhao, Ze Wu, Yueqing Zhuang, Boxun Li, and Jiaya Jia.
\newblock Tracking objects as pixel-wise distributions.
\newblock In {\em Proceedings of the European Conference on Computer Vision},
  pages 76--94. Springer, 2022.

\bibitem{zhou2022slot}
Yi Zhou, Hui Zhang, Hana Lee, Shuyang Sun, Pingjun Li, Yangguang Zhu, ByungIn
  Yoo, Xiaojuan Qi, and Jae-Joon Han.
\newblock Slot-vps: Object-centric representation learning for video panoptic
  segmentation.
\newblock In {\em Proceedings of the IEEE/CVF Conference on Computer Vision and
  Pattern Recognition}, pages 3093--3103, 2022.

\bibitem{zhu2017deep}
Xizhou Zhu, Yuwen Xiong, Jifeng Dai, Lu Yuan, and Yichen Wei.
\newblock Deep feature flow for video recognition.
\newblock In {\em Proceedings of the IEEE/CVF Conference on Computer Vision and
  Pattern Recognition}, 2017.

\bibitem{zhu2019improving}
Yi Zhu, Karan Sapra, Fitsum~A Reda, Kevin~J Shih, Shawn Newsam, Andrew Tao, and
  Bryan Catanzaro.
\newblock Improving semantic segmentation via video propagation and label
  relaxation.
\newblock In {\em Proceedings of the IEEE/CVF Conference on Computer Vision and
  Pattern Recognition}, 2019.

\end{thebibliography}
